\documentclass[conference]{IEEEtran}
\IEEEoverridecommandlockouts
\usepackage{cite}
\usepackage{amsmath,amssymb,amsfonts}
\usepackage{algorithmic}
\usepackage{graphicx}
\usepackage{textcomp}
\usepackage{xcolor}
\usepackage{subfigure}      
\usepackage{url}
\usepackage{amsfonts}
\usepackage{booktabs}       
\usepackage{url}
\usepackage{multicol}
\usepackage{multirow}
\def\BibTeX{{\rm B\kern-.05em{\sc i\kern-.025em b}\kern-.08em
    T\kern-.1667em\lower.7ex\hbox{E}\kern-.125emX}}
\begin{document}

\title{Global Convolutional Neural Processes\\

}
\makeatletter
\newcommand{\linebreakand}{%
  \end{@IEEEauthorhalign}
  \hfill\mbox{}\par
  \mbox{}\hfill\begin{@IEEEauthorhalign}
}
\makeatother
\author{\IEEEauthorblockN{Xuesong Wang}
\IEEEauthorblockA{\textit{University of New South Wales}\\
xuesong.wang1@unsw.edu.au}
\and
\IEEEauthorblockN{Lina Yao}
\IEEEauthorblockA{\textit{University of New South Wales}\\
lina.yao@unsw.edu.au}
\and
\IEEEauthorblockN{Xianzhi Wang}
\IEEEauthorblockA{\textit{University of Technology Sydney}\\
xianzhi.wang@uts.edu.au}
\linebreakand
\IEEEauthorblockN{Hye-young Paik}
\IEEEauthorblockA{\textit{University of New South Wales}\\
h.paik@unsw.edu.au}
\and
\IEEEauthorblockN{Sen Wang}
\IEEEauthorblockA{
\textit{University of Queensland}\\
sen.wang@uq.edu.au}
}

\maketitle

\begin{abstract}
The ability to deal with uncertainty in machine learning models has become equally, if not more, crucial to  their predictive ability itself. For instance, during the pandemic,  governmental policies and personal decisions are constantly made  around uncertainties. Targeting this, Neural Process Families (NPFs) have recently shone a light on prediction with uncertainties by bridging Gaussian processes and neural networks. Latent neural process, a member of NPF, is believed to be capable of modelling the uncertainty on certain points (local uncertainty) as well as the general function priors (global uncertainties). Nonetheless, some critical questions remain unresolved, such as a formal definition of global uncertainties, the causality behind global uncertainties, and the manipulation of global uncertainties for generative models. Regarding this, we build a member GloBal Convolutional Neural Process(GBCoNP) that achieves the SOTA log-likelihood in latent NPFs. It designs a global uncertainty representation \(p(\textbf{\textit{z}})\), which is an aggregation on a discretized input space. The causal effect between the degree of global uncertainty and the intra-task diversity is discussed. The learnt prior is analyzed on a variety of scenarios, including 1D, 2D, and a newly proposed spatial-temporal COVID dataset. Our manipulation of the global uncertainty not only achieves generating the desired samples to tackle few-shot learning, but also enables the probability evaluation on the functional priors.

\end{abstract}


\section{Introduction}
In recent years, machine learning, especially deep learning, has shown massive success on a range of prediction tasks, such as  time-series forecasting~\cite{icdm20stockprediction}, geographical and spatial-temporal inference in medical science, engineering, and finance domains~\cite{icdm20spatialtemporalprediction}. Nonetheless, the uncertainty of machine learning models is less considered than model predictions themselves. When the existing knowledge of a task is not abundant to present a deterministic prediction, uncertainty provides a reasonable guess interval that includes major possibilities of the predictions. In fact, uncertainty can be as equally important as the prediction capability of models, considering that it increases as the expansion of underlying factors.
For example, when estimating the life expectancy of a mechanical part, it is more feasible to estimate an approximate time-range than to predict an exact time stamp at which to discard the part as there are more uncontrollable variants in the long term.
Besides, modelling the uncertainty helps increase models' tolerance to task diversities in datasets.
For instance, when exploring substances from satellite images of a planet, modelling the uncertainty helps incorporate heterogeneous substance candidates that can not be confidently distinguished from the ground truth.
Uncertainty is crucially important for predictions related to the pandemic---it determines how much we could trust the classification results of COVID-19 cases~\cite{chen2020residualUnet}, the anticipated virus spreading trends~\cite{zeroual2020deepcovid}, and assessment of lockdown policies~\cite{qian2020covid}, all of which have a vital impact on the daily lives of many people.

Neural process families (NPF) recently shone a light on predictions with uncertainties by bridging Gaussian processes (GPs) and neural networks. Inherited from Gaussian processes, they make predictions under a joint of correlated normal distributions and present each prediction along with a confidence interval,i.e, uncertainty (See Fig \ref{fig: demo1d}).  As neural networks show competence in deep feature representation, NPFs advance GPs in modelling complicated functions efficiently. 
In addition, NPFs are suitable for solving meta-learning tasks, where each task is sampled from a distribution of functions instead of a single function.
We illustrate predictions with uncertainties using the examples shown in Fig \ref{fig: demo1d}.
Given a \textbf{context set} with a cluster of observable data \((\textbf{\textit{x}}_\mathcal{C}, \textbf{\textit{y}}_\mathcal{C}):=(\textbf{\textit{x}}_i, \textbf{\textit{y}}_i)_{i\in \mathcal{C}}\), an encoder network infers a function \(\textbf{\textit{f}}\) that can generate the context set. Then, a decoder network uses the function to make predictions with uncertainties on a \(\textbf{target set}\) \((\textbf{\textit{x}}_\mathcal{T}, \textbf{\textit{y}}_\mathcal{T}):=(\textbf{\textit{x}}_i, \textbf{\textit{y}}_i)_{i\in \mathcal{T}}\), where only the locations \(\textbf{\textit{x}}_\mathcal{T}\) are unveiled. The resulting outputs follow a normal distribution  \(p(\textbf{\textit{y}}_\mathcal{T}|\textbf{\textit{x}}_\mathcal{T}, \textbf{\textit{f}}) = \mathcal{N}(\mu_\textbf{\textit{y}}, \sigma^2_\textbf{\textit{y}})\). We call the standard deviation on local target points \(\sigma_{\textbf{\textit{y}}_\mathcal{T}}\) ``local uncertainty" in contrast to the global uncertainty to be introduced in latent neural processes later. Eventually, NPF optimizes the parameters in neural nets by maximizing the likelihood of actual target values \(\textbf{\textit{y}}_\mathcal{T}\). 
\begin{figure}[h]
    \centering
        {\includegraphics[width=0.85\linewidth]{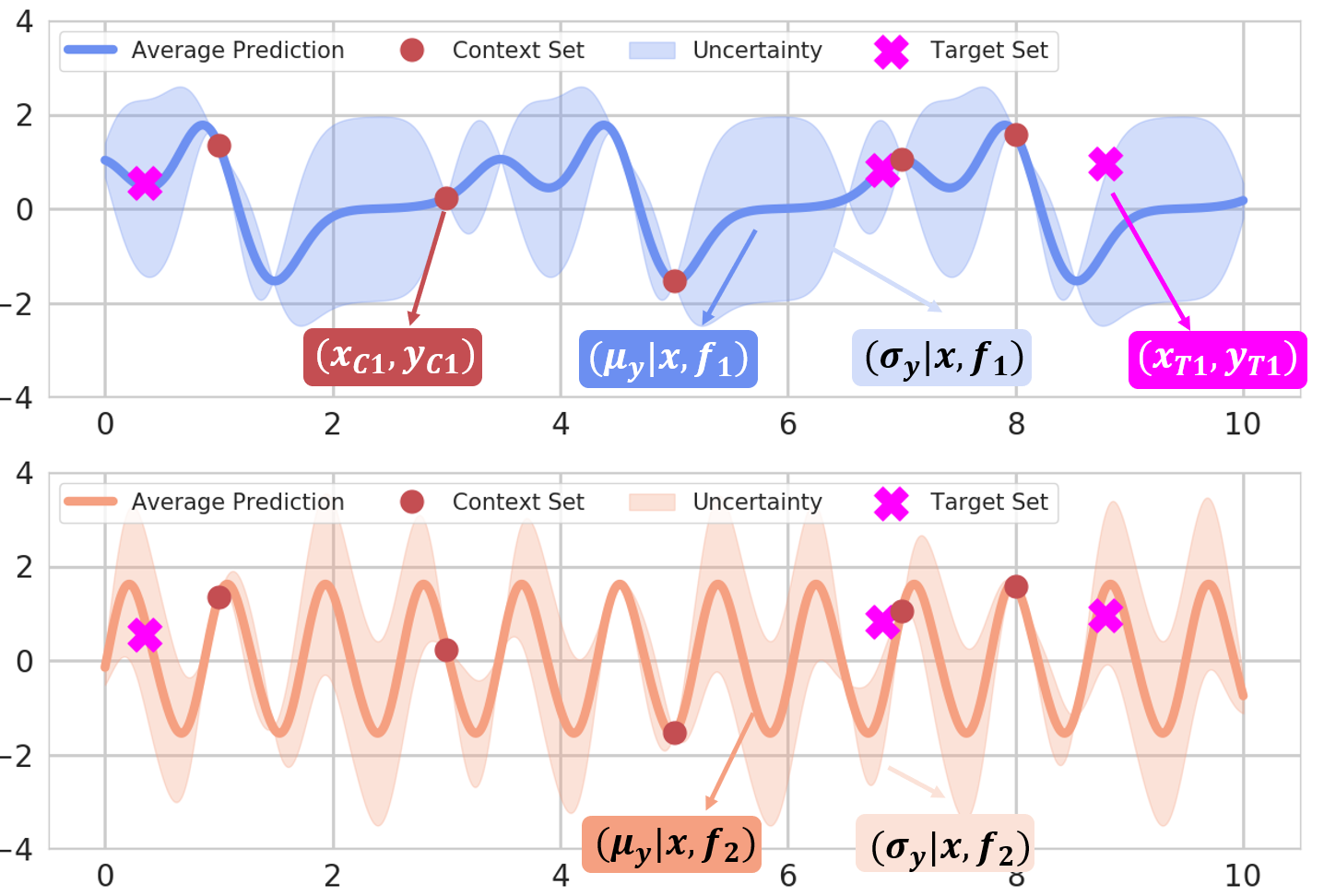}}
    \caption{Predictions with uncertainties under two function samples \(\textbf{\textit{f}}_1\) and \(\textbf{\textit{f}}_2\)}
    \label{fig: demo1d}
\end{figure}

Latent neural processes~\cite{garnelo2018neural} hypothesize that the encoded function should not come from  a deterministic vector but rather a distribution \(\textbf{\textit{f}}\sim p(\textbf{\textit{f}})\). As shown in
Fig \ref{fig: demo1d},
two descent function samples \(\textbf{\textit{f}}_1\) and \(\textbf{\textit{f}}_2\) with periodic differences can be generated using the same context set. They both represent the local uncertainty on the target points but obviously have different priors, meaning there exists another uncertainty that determines this general prior, which we call ``global uncertainty". Despite some previous efforts ~\cite{kim2018attentive}\cite{garnelo2018neural}\cite{foong2020meta_convnp} empirically illustrate such global uncertainties
there remains three unresolved issues:
\begin{itemize} 
\item Formalization of global uncertainties. We wonder if the global uncertainty can be quantified and compared in ways other than empirical visualization of diverse samples.
\item Causality behind global uncertainties. This concerns the model-wise and data-wise factors that affect the global uncertainty. 
\item Manipulation of the global uncertainty for data generation. This concerns tailoring the priors and sampling a desired function once the causal effects of the global uncertainty are examined.
\end{itemize}

We propose a GloBal Convolutional Neural Processes (GBCoNP) that make predictions with uncertainties to address the above challenges. GBCoNP defines global uncertainty as the learnt posterior of a functional distribution conditioned on a small context set \(q(\textbf{\textit{z}}|\mathcal{C})\). This formalization enables comparisons among datasets and different latent models, including the causal-effect between the intra-task diversity and global uncertainty. It further enables us to discover and edit insightful semantic features with regards to global uncertainty during sample generations.
Finally, we evaluate the log-likelihood of GBCoNP with peers on extensive 1D and 2D datasets, and propose a COVID dataset using spatial-temporal uncertainty prediction. This case study is expected to enhance our understanding of the patterns in virus spread and benefit the research community. Our major contributions are three folds:
\begin{itemize} 
\item A new discretized space for global uncertainty projection that is suitable for out-of-range prediction while maintaining the shared global prior.
\item A causal-effect analysis of the global uncertainty, which is seldom discussed in previous research. Our analysis reveals dataset characteristics, such as intra-task diversity, can depict the stochasticity.
\item Manipulation of the global uncertainty that empowers sampling with priors. We novelly generalize the applications to a high-dimensional spatial-temporal scenario.
\end{itemize}

\section{Methodology}
 
\subsection{Global Uncertainty in Latent Neural Processes}
We follow the notations of \cite{kim2018attentive} and denote a context set by \((\textbf{\textit{x}}_\mathcal{C}, \textbf{\textit{y}}_\mathcal{C}):=(\textbf{\textit{x}}_i, \textbf{\textit{y}}_i)_{i\in \mathcal{C}}\), where both the inputs and predictions are given in a meta-regression task.
The context set defines a sample from a functional distribution. The objective is to predict on the target inputs \(\textbf{\textit{x}}_\mathcal{T}\) using this function sample and maximize the likelihood of \(\textbf{\textit{y}}_\mathcal{T}\) if the target set \((\textbf{\textit{x}}_\mathcal{T}, \textbf{\textit{y}}_\mathcal{T}):=(\textbf{\textit{x}}_i, \textbf{\textit{y}}_i)_{i\in \mathcal{T}}\) comes from the same function sample. Similar to conditional variational autoencoders (CVAE), a latent NP models the prediction with a conditional distribution \eqref{equ:dist of p_y}:
\begin{equation}
\begin{gathered}
    p(\textbf{\textit{y}}_\mathcal{T} | \textbf{\textit{x}}_\mathcal{T}, \textbf{\textit{x}}_\mathcal{C}, \textbf{\textit{y}}_\mathcal{C}):= \mathcal{N}(\mu_\textbf{\textit{y}},  \sigma^2_\textbf{\textit{y}}) \\
    =\int p(\textbf{\textit{y}}_\mathcal{T} | \textbf{\textit{x}}_\mathcal{T}, \textbf{\textit{r}}_\mathcal{C},\textbf{\textit{z}})p(\textbf{\textit{z}})\,d\textbf{\textit{z}}
\end{gathered}
\label{equ:dist of p_y}
\end{equation}
where \(\textbf{\textit{r}}_\mathcal{C}:= r(\textbf{\textit{x}}_\mathcal{C}, \textbf{\textit{y}}_\mathcal{C})\) is a neural network that captures the deterministic part of the functional sample.
The prior latent distribution \(p(\textbf{\textit{z}}):=\mathcal{N}(\mu_\textbf{\textit{z}},  \sigma^2_\textbf{\textit{z}})\) captures the global uncertainty of the functional sample.
As shown in Fig \ref{fig: workflow of NP and ConvCNP}(a),  \(\textbf{\textit{r}}_\mathcal{C}\) and \(\textbf{\textit{z}}\) forms the encoder. The decoder network takes the encoded functional condition \((\textbf{\textit{r}}_\mathcal{C}, \textbf{\textit{z}})\) and the target inputs \(\textbf{\textit{x}}_\mathcal{T}\) to make predictions of \(\textbf{\textit{y}}_\mathcal{T} \). Depending on the inductive biases imposed over the relationships within the context set and between context and target sets, different neural network structures can be adopted for encoders, such as equally weighted (neural processes \cite{garnelo2018neural}) and attentively weighted aggregation (attentive neural processes \cite{kim2018attentive}).

\begin{figure}[!b]
    \centering
    \begin{subfigure}[Latent NP]
        {\includegraphics[width=0.8\linewidth]{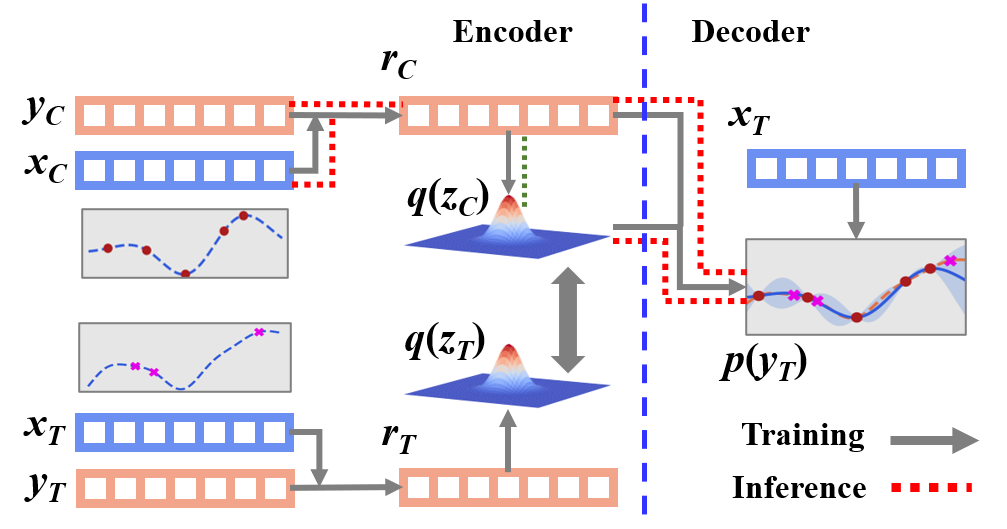}}
    \end{subfigure}
    \begin{subfigure}[ConvCNP]
         {\includegraphics[width=0.8\linewidth]{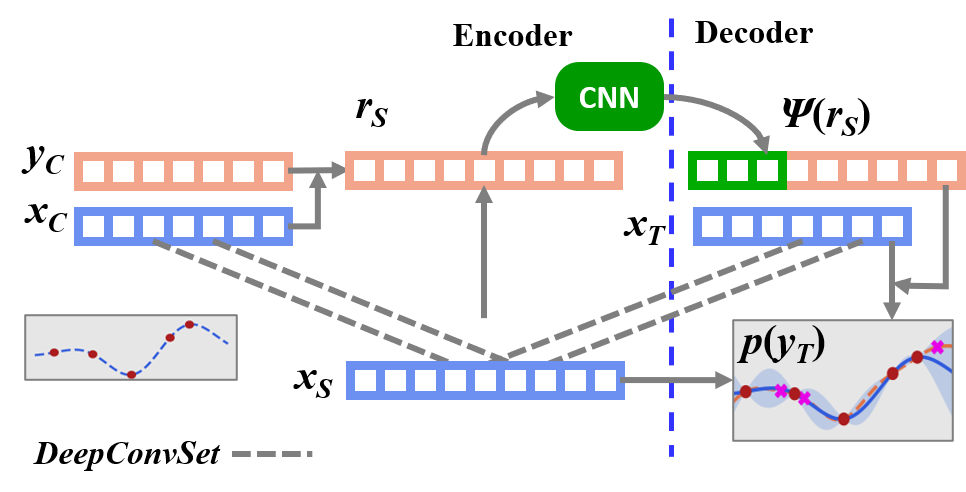}}
    \end{subfigure}
    \caption{Workflows for latent NP families and ConvCNP}
    \label{fig: workflow of NP and ConvCNP}
\end{figure}

Since this prior knowledge of the functional distribution, i.e., global uncertainty \(p(\textbf{\textit{z}})\) is intractable, previous studies turn to amortised variational posterior \(p(\textbf{\textit{z}}) = \mathcal{N}(\mu_\textbf{\textit{z}}, \sigma^2_\textbf{\textit{z}}) \approx  q(\textbf{\textit{z}}| \textbf{\textit{r}}_\mathcal{C})=q(\textbf{\textit{z}}| \textbf{\textit{x}}_\mathcal{C}, \textbf{\textit{y}}_\mathcal{C})\) for inference.
They typically  pass \(\textbf{\textit{r}}_\mathcal{C}\) through an MLP to get the distributional parameters and then optimize \eqref{equ:dist of p_y} with a differentiable network:
\begin{equation}
\begin{gathered}
    p(\textbf{\textit{y}}_\mathcal{T} | \textbf{\textit{x}}_\mathcal{T}, \textbf{\textit{x}}_\mathcal{C}, \textbf{\textit{y}}_\mathcal{C}):= \int p(\textbf{\textit{y}}_\mathcal{T} | \textbf{\textit{x}}_\mathcal{T}, \textbf{\textit{r}}_\mathcal{C},\textbf{\textit{z}})q(\textbf{\textit{z}}| \textbf{\textit{x}}_\mathcal{C}, \textbf{\textit{y}}_\mathcal{C}))\,d\textbf{\textit{z}}
\end{gathered}
\label{equ: posterior of p_y}
\end{equation}
A predictive evidence-lower-bound (ELBO) is given based on the variational inference of \(p(\textbf{\textit{z}})\):
\begin{equation}
\begin{gathered}
    \log p(\textbf{\textit{y}}_\mathcal{T} | \textbf{\textit{x}}_\mathcal{T}, \textbf{\textit{x}}_\mathcal{C}, \textbf{\textit{y}}_\mathcal{C}) \geq  \mathbb{E}_{q(\textbf{\textit{z}}|\textbf{\textit{x}}_\mathcal{T}, \textbf{\textit{y}}_\mathcal{T})}{\log p(\textbf{\textit{y}}_\mathcal{T} | \textbf{\textit{x}}_\mathcal{T}, \textbf{\textit{r}}_\mathcal{C},\textbf{\textit{z}})}\\ - \mathcal{KL}(q(\textbf{\textit{z}}|\textbf{\textit{x}}_\mathcal{T}, \textbf{\textit{y}}_\mathcal{T})||q(\textbf{\textit{z}}|\textbf{\textit{x}}_\mathcal{C}, \textbf{\textit{y}}_\mathcal{C}))
\end{gathered}
\label{equ: ELBO}
\end{equation}

Now, the training process is to maximize the log-likelihood \(\log p(\textbf{\textit{y}}_\mathcal{T} | \textbf{\textit{x}}_\mathcal{T}, \textbf{\textit{x}}_\mathcal{C}, \textbf{\textit{y}}_\mathcal{C})\) which equals maximizing its lower bound comprised of a conditional log-likelihood based on latent z, i.e., \(\log p(\textbf{\textit{y}}_\mathcal{T} | \textbf{\textit{x}}_\mathcal{T}, \textbf{\textit{r}}_\mathcal{C},\textbf{\textit{z}})\), minus a non-negative KL divergence.
While CVAEs regularize the posterior with a standard normal distribution \(\mathcal{KL}({q(\textbf{\textit{z}})|| \mathcal{N}(0, I)})\), latent NPs differ in minimizing the divergence from the posterior (obtained from the target set) to the context set during training when the target sets are accessible. 

During inference, the prior \(p(\textbf{\textit{z}}) = \mathcal{N}(\mu_\textbf{\textit{z}}, \sigma^2_\textbf{\textit{z}})\) is replaced by \( q(\textbf{\textit{z}}| \textbf{\textit{x}}_\mathcal{C}, \textbf{\textit{y}}_\mathcal{C})\). We formalize the global uncertainty as:
\begin{equation}
\begin{gathered}
    p(\textbf{\textit{z}}):=\mathcal{N}(\mu_\textbf{\textit{z}}, \sigma^2_\textbf{\textit{z}}) = q(\textbf{\textit{z}}|\mathcal{C})\), \(s.t. |\mathcal{C}|< \epsilon
\end{gathered}
\label{equ: global uncertainty}
\end{equation}

where \(\epsilon\) comprises a very small proportion of the index set. We did not set the condition \(\mathcal{C} = \varnothing\) since in real-world datasets and the empty set barely carries any function prior, but \(|\mathcal{C}|\) cannot be too large, otherwise it would leave no space for uncertainty. The mean \(\mu_\textbf{\textit{z}}\) in \eqref{equ: global uncertainty} determines the sensitivity of final predictions affected by the global uncertainty, whereas the variance \(\sigma^2_\textbf{\textit{z}}\) implies the diversity of function priors on a certain task.

\subsection{Out-of-range Predictions with Convolution}
Out-of-range predictions is an essential characteristic when scaling the NPF to real-world tasks. It requires the model to generalize predictions when the testing task is out of the training range. The recently proposed Convolutional Conditional Neural Processes(ConvCNP)\cite{gordon2019convolutional} tackle this issue with convolutions and outperform peer NPF members. ConvCNPs assume ``translational invariance", i.e., the prediction pattern near a local context data is transferable to the rest of the input space, which can be well addressed by convolution. They omit the latent representation and only decode \(\textbf{\textit{r}}\) and \(\textbf{\textit{x}}_\mathcal{T}\). However, instead of directly encoding \(\textbf{\textit{r}}\) from the raw context inputs \(\textbf{\textit{x}}_\mathcal{C}\),  they introduce a discretization of the indefinite input space \(\textbf{\textit{x}}_\mathcal{S}\) that incorporates context and target inputs \(\textbf{\textit{x}}_\mathcal{C}, \textbf{\textit{x}}_\mathcal{T}\) and map the deterministic representation on \(\textbf{\textit{x}}_\textbf{\textit{S}}\):
\begin{equation}
\begin{gathered}
    p(\textbf{\textit{y}}_\mathcal{T} | \textbf{\textit{x}}_\mathcal{T}, \textbf{\textit{x}}_\mathcal{C}, \textbf{\textit{y}}_\mathcal{C}):= \mathcal{N}(\mu_\textbf{\textit{y}},  \sigma^2_\textbf{\textit{y}})=
    p(\textbf{\textit{y}}_\mathcal{T} | \textbf{\textit{x}}_\mathcal{T}, \psi(\textbf{\textit{r}}_\mathcal{S}))
\end{gathered}
\label{equ: ConvCNP: r_S}
\end{equation}
where \(\textbf{\textit{r}}_\mathcal{S}:= r(\textbf{\textit{x}}_\mathcal{C},\textbf{\textit{x}}_\mathcal{S}, \textbf{\textit{y}}_\mathcal{C})\) is the deterministic representation. \(r(\textbf{\textit{x}}_\mathcal{C},\textbf{\textit{x}}_\mathcal{S}, \textbf{\textit{y}}_\mathcal{C})\) is a \(\textit{DeepConvSet}\) that resembles a simplified attention.
Given a query of a discretized set \(\textbf{\textit{x}}_\mathcal{S}\), \(\textit{DeepConvSet}\) projects the value \(\textbf{\textit{y}}_\mathcal{C}\) to the space \(\mathcal{S}\) based on the similarity between the query \(\textbf{\textit{x}}_\mathcal{S}\) and the key \(\textbf{\textit{x}}_\mathcal{C}\).
Then, \(\textbf{\textit{r}}_\mathcal{S}\) is passed through a convolutional neural network \(\psi(\cdot)\), which achieves the inductive bias of transitional invariance (shown in Fig \ref{fig: workflow of NP and ConvCNP}(b)).
This guarantees that the function description only focuses on a local range of inputs so that the model is still effective even when the testing inputs are out of the training range.
The \(\textit{DeepConvSet}:\)  \(\mathcal{C}\mapsto\mathcal{S}\) and the CNN module constitute the encoder. The decoder is another \(\textit{DeepConvSet}\) whose keys and values are \(\textbf{\textit{x}}_\mathcal{S}\) and \(\textbf{\textit{r}}_\mathcal{S}\) with queries \(\textbf{\textit{x}}_\mathcal{T}\). Without the latent representation, the training and inference share the same workflow.

\begin{figure}[!b]
    \centering
    {\includegraphics[width=0.9\linewidth]{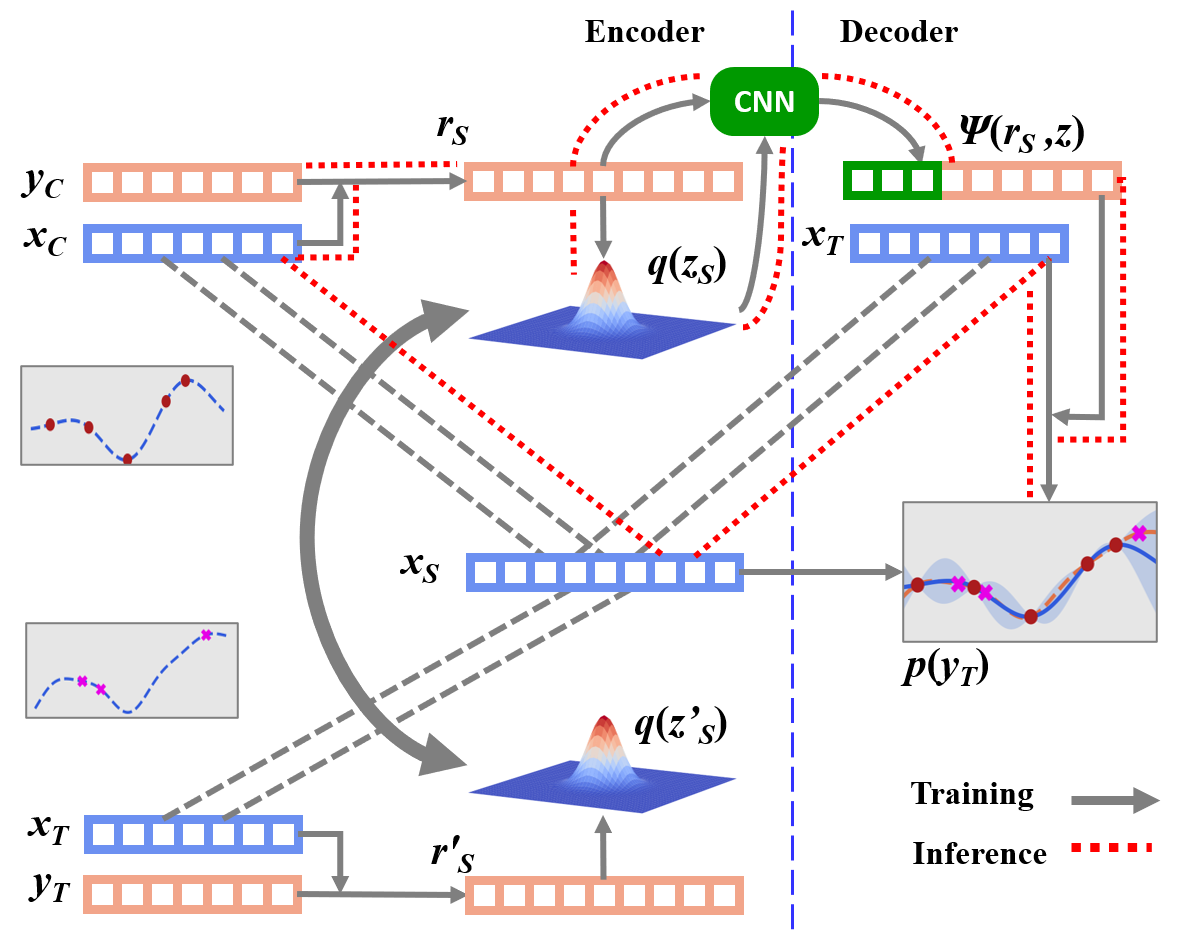}}
    \caption{The model structure for GBCoNP}
    \label{fig:workflow for GBCoNP}
\end{figure}

\subsection{GloBal Convolutional Neural Process(GBCoNP)}
    While ConvCNPs outperform NPF members on many scenarios, we believe the latent distribution \(p(\textbf{\textit{z}})\) has a great impact on maintaining the stochasticity for NPs, particularly on the global uncertainty. If the function sample representation \(\textbf{\textit{r}}_\mathcal{S}\) is deterministic, the resulting prediction will be precise yet ``dull" with single distribution parameters \(\mu_\textbf{\textit{y}}\) and \(\sigma^2_\textbf{\textit{y}}\); In contrast, latent NPs can sample different \(\textbf{\textit{z}}\) values which correspond to diverse priors over the functions. Each prior is able to generate a cluster of \((\mu_\textbf{\textit{y}}, \sigma^2_\textbf{\textit{y}})\). ConvCNP can potentially be tailored to a latent NP, given that the mapping function to the space \(\mathcal{S}\) can be latent and shared between the context and target set. Therefore, we introduce a member named GloBal Convolutional Neural Process (GBCoNP) that also adopts amortised variational inference on the global uncertainty (shown in Fig \ref{fig:workflow for GBCoNP}).


The predictions of GBCoNP follow a conditional distribution with a latent path:
\begin{equation}
\begin{gathered}
    p(\textbf{\textit{y}}_\mathcal{T} | \textbf{\textit{x}}_\mathcal{T}, \textbf{\textit{x}}_\mathcal{C}, \textbf{\textit{y}}_\mathcal{C}):= \mathcal{N}(\mu_\textbf{\textit{y}},  \sigma^2_\textbf{\textit{y}}) \\
    =\int p(\textbf{\textit{y}}_\mathcal{T} | \textbf{\textit{x}}_\mathcal{T}, \psi(\textbf{\textit{r}}_\mathcal{S},\textbf{\textit{z}}))p(\textbf{\textit{z}})\,d\textbf{\textit{z}}
\end{gathered}
\label{equ:dist of GBCoNP p_y}
\end{equation}
where \(\psi(\cdot)\) is the convolution module applied on both the deterministic and the latent representation. \(\textbf{\textit{r}}_\mathcal{S}:=r(\textbf{\textit{x}}_\mathcal{C},\textbf{\textit{x}}_\mathcal{S}, \textbf{\textit{y}}_\mathcal{C})\) is a \(\textit{DeepConvSet}\). Given the convolved functional condition \(\psi(\textbf{\textit{r}}_\mathcal{S},\textbf{\textit{z}})\) and the target inputs \(\textbf{\textit{x}}_\mathcal{T}\), a decoder \(\textit{DeepConvSet}\) maps the condition to the prediction distribution.
We use a new space \(\textbf{\textit{x}}_\mathcal{S}\) instead of the original space \(\textbf{\textit{x}}_\mathcal{C}\) to obtain the variational inference of the intractable prior \(p(\textbf{\textit{z}})\). Thus, \eqref{equ:dist of GBCoNP p_y} can be optimized with another differentiable network (\(\psi(\cdot)\) omitted for better clarification):
\begin{equation}
\begin{gathered}
    p(\textbf{\textit{y}}_\mathcal{T} |\textbf{\textit{x}}_\mathcal{T}, \textbf{\textit{x}}_\mathcal{C}, \textbf{\textit{y}}_\mathcal{C}):=\int p(\textbf{\textit{y}}_\mathcal{T} |\textbf{\textit{x}}_\mathcal{T}, \textbf{\textit{r}}_\mathcal{S},\textbf{\textit{z}}) q(\textbf{\textit{z}}|\textbf{\textit{r}}_\mathcal{S})\,d\textbf{\textit{z}}
\end{gathered}
\label{equ:variation inference of GBCoNP p_y}
\end{equation}

According to \cite{garnelo2018neural}, the joint distribution of \(p(\textbf{\textit{y}}_\mathcal{T})\) must suffice exchangeability to be a stochastic process:
\begin{equation}
\begin{gathered}
    p(\textbf{\textit{y}}_{\mathcal{T}_1}, \textbf{\textit{y}}_{\mathcal{T}},
    ..., \textbf{\textit{y}}_{\mathcal{T}_n}|
    \textbf{\textit{x}}_{\mathcal{T}_1},
    ..., \textbf{\textit{x}}_{\mathcal{T}_n},
    \textbf{\textit{x}}_{\mathcal{C}_1},
    ..., \textbf{\textit{x}}_{\mathcal{C}_m},
    \textbf{\textit{y}}_{\mathcal{C}_1},
    ..., \textbf{\textit{y}}_{\mathcal{C}_m})\\
    =  p(\textbf{\textit{y}}_{\mathcal{T}_{\pi(1)}},
    ..., \textbf{\textit{y}}_{\mathcal{T}_{\pi(n)}}|
    \textbf{\textit{x}}_{\mathcal{T}_{\pi(1)}},
    ..., \textbf{\textit{x}}_{\mathcal{T}_{\pi(n)}},\\    \textbf{\textit{x}}_{\mathcal{C}_{\pi'(1)}},
    ..., \textbf{\textit{x}}_{\mathcal{C}_{\pi'(m)}},
    \textbf{\textit{y}}_{\mathcal{C}_{\pi'(1)}},
    ..., \textbf{\textit{y}}_{\mathcal{C}_{\pi'(m)}})
\end{gathered}
\label{equ:permutation invariance}
\end{equation}

where \(\pi\) and \(\pi'\) are permutations of the target index set \(\{1, ..., n\}\) and the context index set \(\{1, ..., m\}\),
meaning the permutation of the context and the target points cannot change the prediction outcome.
The previous latent NPs need an \textit{\textbf{aggregation module}} (mean, sum, etc.) on context and target space to ensure \(q(\textbf{\textit{z}}|\textbf{\textit{x}}_\mathcal{T}, \textbf{\textit{y}}_\mathcal{T})\) and \(q(\textbf{\textit{z}}|\textbf{\textit{x}}_\mathcal{C}, \textbf{\textit{y}}_\mathcal{C})\) are permutation invariant and have the identical dimensions for divergence. 

Similar to attention, the \(\textit{DeepConvSet}\) already suffices the exchangeability by calculating the inner product between the key (\(\textbf{\textit{x}}_\mathcal{C}\) or \(\textbf{\textit{x}}_\mathcal{T}\)) and query \(\textbf{\textit{x}}_\mathcal{S}\). This operation is insensitive to the input order and thus the context and target are projected to the same space \(\mathcal{S}\) regardless of the order. Presumably, \(\textbf{\textit{r}}_\mathcal{S}\) can be directly passed through an MLP for \(\textbf{\textit{z}}\) without aggregation. However, we discovered that the aggregation on \(\textbf{\textit{z}}\) can mitigate the coherence deficiency in the samples, a drawback caused by the independent prediction assumption in NPFs\cite{foong2020meta_convnp}. It maybe attributed to the diminishing effect on the divergence between \(q(\textbf{\textit{z}}|\textbf{\textit{r}}_\mathcal{S}):=q(\textbf{\textit{z}}|\textbf{\textit{x}}_\mathcal{S},\textbf{\textit{x}}_\mathcal{C}, \textbf{\textit{y}}_\mathcal{C})\) and \(q(\textbf{\textit{z}}|\textbf{\textit{r}}_\mathcal{S}'):=q(\textbf{\textit{z}}|\textbf{\textit{x}}_\mathcal{S},\textbf{\textit{x}}_\mathcal{T}, \textbf{\textit{y}}_\mathcal{T})\) after the fusion. 

An observation that supports the aggregation is the non-translational invariance of the pre-aggregated \(\textbf{\textit{z}}\). Normally, \(\psi(\cdot)\) in \eqref{equ:dist of GBCoNP p_y} comprises several cascaded convolutional nets; when we directly convolve on \(\textbf{\textit{r}}_\mathcal{S}\) concatenated with the pre-aggregated \(\textbf{\textit{z}}\), the local pattern near a context point (e.g., a fluctuation) is propagated to the entire space \(\mathcal{S}\). Besides, similar patterns are obtained when \(\textbf{\textit{z}}\) is derived from an intermediate convolution state of \(\psi(\textbf{\textit{r}}_\mathcal{S})\) and get aggregated afterwards, meaning \(\textbf{\textit{r}}_\mathcal{S}\) is convolved by a subset of the cascaded nets. Both cases above imply that local patterns on \(\textbf{\textit{z}}\) cannot be as transferable to the whole space 
as \(\textbf{\textit{r}}_\mathcal{S}\) can. Therefore, we add an aggregation module to \(\textbf{\textit{z}}\) to ensure every point in the space \(\mathcal{S}\) gets the same latent prior. 

Adopting the variation inference, the objective function for GBCoNP now becomes \eqref{equ: ELBO for GBCoNP}:

\noindent\textbf{Prop 1.} Evidence-Lower-BOund for GBCoNP: 
\begin{equation}
\begin{gathered}
    \log p(\textbf{\textit{y}}_\mathcal{T} | \textbf{\textit{x}}_\mathcal{T}, \textbf{\textit{x}}_\mathcal{C}, \textbf{\textit{y}}_\mathcal{C})
    \geq \mathbb{E}_{q(\textbf{\textit{z}}|\textbf{\textit{r}}_\mathcal{S}')}{\log p(\textbf{\textit{y}}_\mathcal{T}| \textbf{\textit{x}}_\mathcal{T}, \textbf{\textit{r}}_\mathcal{S}, \textbf{\textit{z}})} \\
    - \mathcal{KL}(q(\textbf{\textit{z}}|\textbf{\textit{r}}_\mathcal{S}')||q(\textbf{\textit{z}}|\textbf{\textit{r}}_\mathcal{S}))
\end{gathered}
\label{equ: ELBO for GBCoNP}
\end{equation}

\noindent\textit{Proof.} The conditional probability of the prediction can be built on the marginalization of a joint distribution with a latent variable z.
\begin{equation}
\begin{gathered}
    \log p(\textbf{\textit{y}}_\mathcal{T} | \textbf{\textit{x}}_\mathcal{T}, \textbf{\textit{x}}_\mathcal{C}, \textbf{\textit{y}}_\mathcal{C})
    = \log\int p(\textbf{\textit{y}}_\mathcal{T}, \textbf{\textit{z}}| \textbf{\textit{x}}_\mathcal{T}, \textbf{\textit{r}}_\mathcal{S})\,d\textbf{\textit{z}}\\
    =\log\int p(\textbf{\textit{y}}_\mathcal{T}| \textbf{\textit{x}}_\mathcal{T}, \textbf{\textit{r}}_\mathcal{S}, \textbf{\textit{z}})p(\textbf{\textit{z}})\,d\textbf{\textit{z}}\\
    =\log\int \frac{p(\textbf{\textit{y}}_\mathcal{T}| \textbf{\textit{x}}_\mathcal{T}, \textbf{\textit{r}}_\mathcal{S}, \textbf{\textit{z}})p(\textbf{\textit{z}})}{q(\textbf{\textit{z}}|\textbf{\textit{x}}_\mathcal{T}, \textbf{\textit{y}}_\mathcal{T}, \textbf{\textit{x}}_\mathcal{S})}q(\textbf{\textit{z}}|\textbf{\textit{x}}_\mathcal{T}, \textbf{\textit{y}}_\mathcal{T}, \textbf{\textit{x}}_\mathcal{S})\,d\textbf{\textit{z}}\\
    = \log\mathbb{E}_{q(\textbf{\textit{z}}|\textbf{\textit{r}}_\mathcal{S}')}\frac{p(\textbf{\textit{y}}_\mathcal{T}| \textbf{\textit{x}}_\mathcal{T}, \textbf{\textit{r}}_\mathcal{S}, \textbf{\textit{z}})p(\textbf{\textit{z}})}{q(\textbf{\textit{z}}|\textbf{\textit{r}}_\mathcal{S}')}\\
\end{gathered}
\label{eq: prior inequility}
\end{equation}
Since the prior \(\textbf{\textit{z}}\) is intractable, it is replaced by the conditional posterior \(q(\textbf{\textit{z}}|\textbf{\textit{r}}_\mathcal{S})\) on the space \(\mathcal{S}\). Using Jensen’s Inequality and the concave \(\log\) function, we get the lower bound of \eqref{eq: prior inequility}:
\begin{equation}
\begin{gathered}
    \geq \mathbb{E}_{q(\textbf{\textit{z}}|\textbf{\textit{r}}_\mathcal{S}')}{\log p(\textbf{\textit{y}}_\mathcal{T}| \textbf{\textit{x}}_\mathcal{T}, \textbf{\textit{r}}_\mathcal{S}, \textbf{\textit{z})}} + \mathbb{E}_{q(\textbf{\textit{z}}|\textbf{\textit{r}}_\mathcal{S}')}\log \frac{q(\textbf{\textit{z}}|\textbf{\textit{r}}_\mathcal{S})}{q(\textbf{\textit{z}}|\textbf{\textit{r}}_\mathcal{S}')} \\
    = \mathbb{E}_{q(\textbf{\textit{z}}|\textbf{\textit{r}}_\mathcal{S}')}{\log p(\textbf{\textit{y}}_\mathcal{T}| \textbf{\textit{x}}_\mathcal{T}, \textbf{\textit{r}}_\mathcal{S}, \textbf{\textit{z}})} -  \mathcal{KL}(q(\textbf{\textit{z}}|\textbf{\textit{r}}_\mathcal{S}')||q(\textbf{\textit{z}}|\textbf{\textit{r}}_\mathcal{S}))
\end{gathered}
\label{eq: post inequility}
\end{equation}

\textbf{Computational complexity.}
An attention layer takes \(\mathcal{O}(nmd)\), where \(n\) and \(m\) are the key and query lengths, and \(d\) implies the weight dimension~\cite{vaswani2017attentionisall}.
A convolutional layer costs \(\mathcal{O}(nkdf)\), where \(k\) and \(f\) refer to the kernel size and channel depth.
Therefore, two \(\textit{DeepConvSet}\)s in encoder and decoder cost \(\mathcal{O}(msd)\) and \(\mathcal{O}(nsd)\), where \(m\) and \(n\) refer to the context and target set sizes, and \(s\) refers to the grid length of the discretization \((s \gg m,n)\).
The computationally expensive part \(\psi(\cdot)\) costs \(\mathcal{O}(skdf)\) for convolutions on the \(\mathcal{S}\) space.
Compared with ConvCNP, the latent path in GBCoNP brought in two MLP modules: the latent MLP from \(\textbf{\textit{r}}_\mathcal{S}\) to \(\textbf{\textit{z}}\) and the merger MLP (which compress the concatenation of \((\textbf{\textit{r}}_\mathcal{S}, \textbf{\textit{z}})\) to a low dimension before convolution). Both modules cost \(\mathcal{O}(s)\). Such modifications preserve the total number of parameters in convolution and are thus affordable.


\section{Experiments}
We evaluate our proposed model on three groups of datasets covering 1D, 2D and spatial temporal scenarios across a broad range of domains. All the models are built with Python 3.6.8 and Pytorch 1.4.0. Two TITAN RTX GPUs are used for training. Wall-clock time for training one epoch of GBCoNP and ConvCNP is shown in Appendix \ref{appendix: running time costs}.
\subsection{1D Datasets}
In 1D scenarios, each task deals with a series of context points. The objective is to predict the values and uncertainty for the unseen target set (The datasets are detailed in Appendix \ref{appendix:Dataset details}). For each task, x values are normalized to [-1, 1] and y values are standard normalized before training. The number of context points range within \(\mathcal{U}(1, 50)\), and the target set comprises all the data points in the task. Each dataset is trained for 100 epochs and tested for 6 runs.

Table \ref{tab1d} shows the log-likelihood of the latent neural process families along with their conditional members on 1D datasets. The poor results of a neural network (NN) that abandons the context set \(y_T =f(x_T)\) reflect that non-NP models are probably unsuitable for the meta-setting. ConvCNP and ANP achieve the best baseline log-likelihood in their category; their predictions are compared further with the proposed GBCoNP in Fig \ref{fig: 1d results}. 

\begin{table*}[h]
\caption{Loglikelihood on 1D Datasets (Mean \(\pm\) Std)}
\begin{center}
\begin{tabular}{|c|c|c|c|c|c|c|}
\hline
\textbf{Model}&\multicolumn{6}{|c|}{\textbf{Dataset Name}} \\
\cline{2-7} 
\textbf{Name} & \textbf{\textit{RBF}}& \textbf{\textit{Periodic}} & \textbf{\textit{Matérn-3/2}}&  \textbf{\textit{Stock50}} & \textbf{\textit{SmartMeter}} & \textbf{\textit{HousePricing}} \\ \hline 
NN      & -1.42 \(\pm\) 0.00 & -1.42 \(\pm\) 0.00 & -1.42 \(\pm\) 0.00 & -1.40 \(\pm\) 4E-3 & -1.42 \(\pm\) 7E-5 & -1.16 \(\pm\) 4E-3   \\
CNP~\cite{garnelo2018conditional}     & -0.73 \(\pm\) 0.03 & -1.27 \(\pm\) 2E-3 & -0.89  \(\pm\) 8E-3 & -0.53  \(\pm\) 0.05 & -0.43  \(\pm\) 0.03 & 0.33  \(\pm\) 0.08    \\
ACNP    & 0.74  \(\pm\) 0.12  & -1.08 \(\pm\)  7E-3 & 0.23  \(\pm\) 0.03  & -0.11  \(\pm\) 0.11 & -0.33  \(\pm\) 0.05 & 1.07  \(\pm\) 0.15    \\
ConvCNP~\cite{gordon2019convolutional} & \textbf{1.19  \(\pm\) 0.15}  &\textbf{ 0.71  \(\pm\) 0.03}  &\textbf{ 0.41  \(\pm\) 0.04}  & \textbf{0.03  \(\pm\) 0.16 } &\textbf{ 0.09  \(\pm\) 0.07}  & \textbf{1.25  \(\pm\) 0.18 }  \\
\cline{1-7} 
NP \cite{garnelo2018neural}     & -1.08  \(\pm\) 0.03 & -1.30  \(\pm\) 2E-3 & -1.04 \(\pm\)  0.01 & -0.79  \(\pm\) 0.08 & -0.47  \(\pm\) 0.02 & -1.20  \(\pm\) 0.39   \\
ANP \cite{kim2018attentive}    & 0.74  \(\pm\) 0.12  & -1.06 \(\pm\)  8E-3 & 0.19 \(\pm\)  0.03  & -0.19  \(\pm\) 0.15 & -0.33  \(\pm\) 0.05 & 0.80  \(\pm\) 0.37    \\
GBCoNP  & \textbf{1.24  \(\pm\) 0.14}  & \textbf{0.25 \(\pm\)  0.04 } & \textbf{0.37 \(\pm\)  0.03 } & \textbf{0.03 \(\pm\)  0.13}  & \textbf{0.02  \(\pm\) 0.05} & \textbf{1.11  \(\pm\) 0.24}    \\ \hline
\end{tabular}
\label{tab1d}
\end{center}
\end{table*}

\begin{figure*}[h!]
    \centering
        {\includegraphics[width=0.32\linewidth]{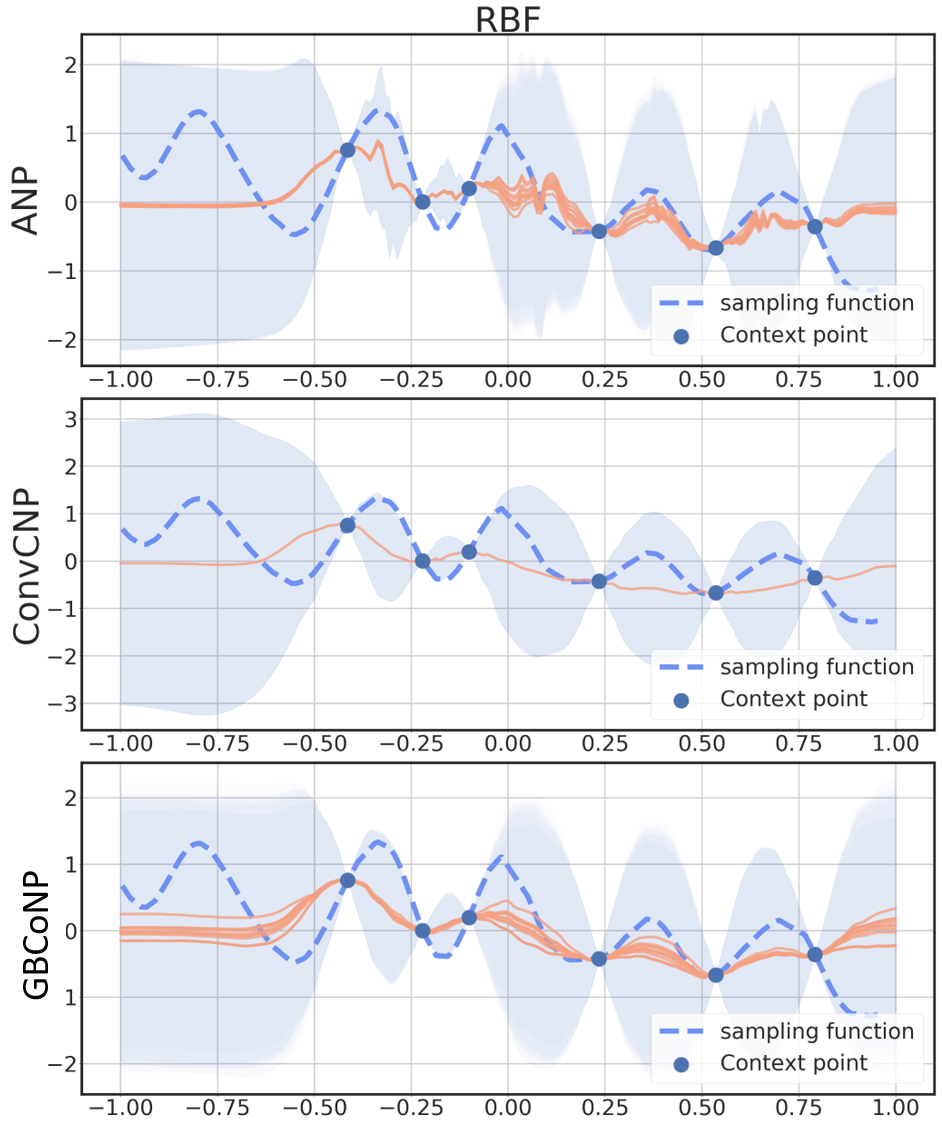}}
        {\includegraphics[width=0.32\linewidth]{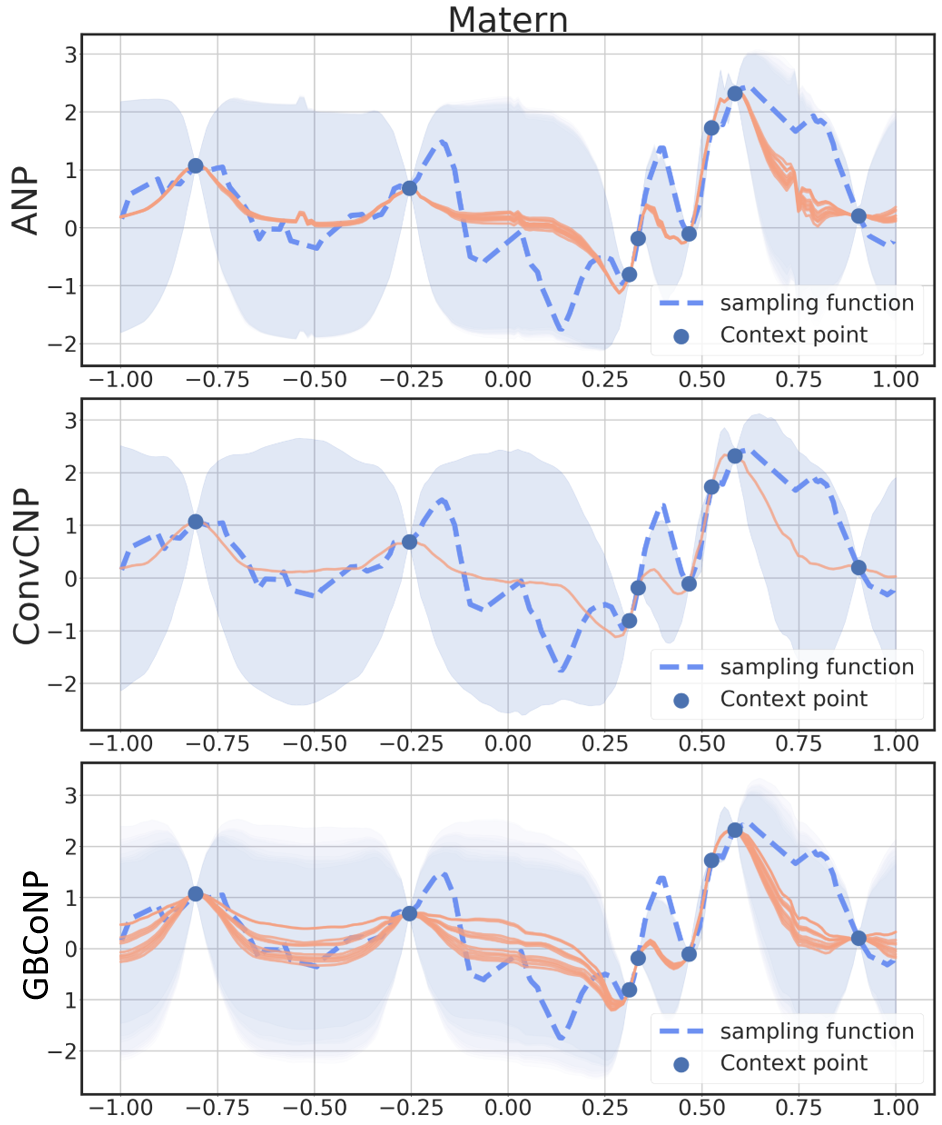}}
        {\includegraphics[width=0.32\linewidth]{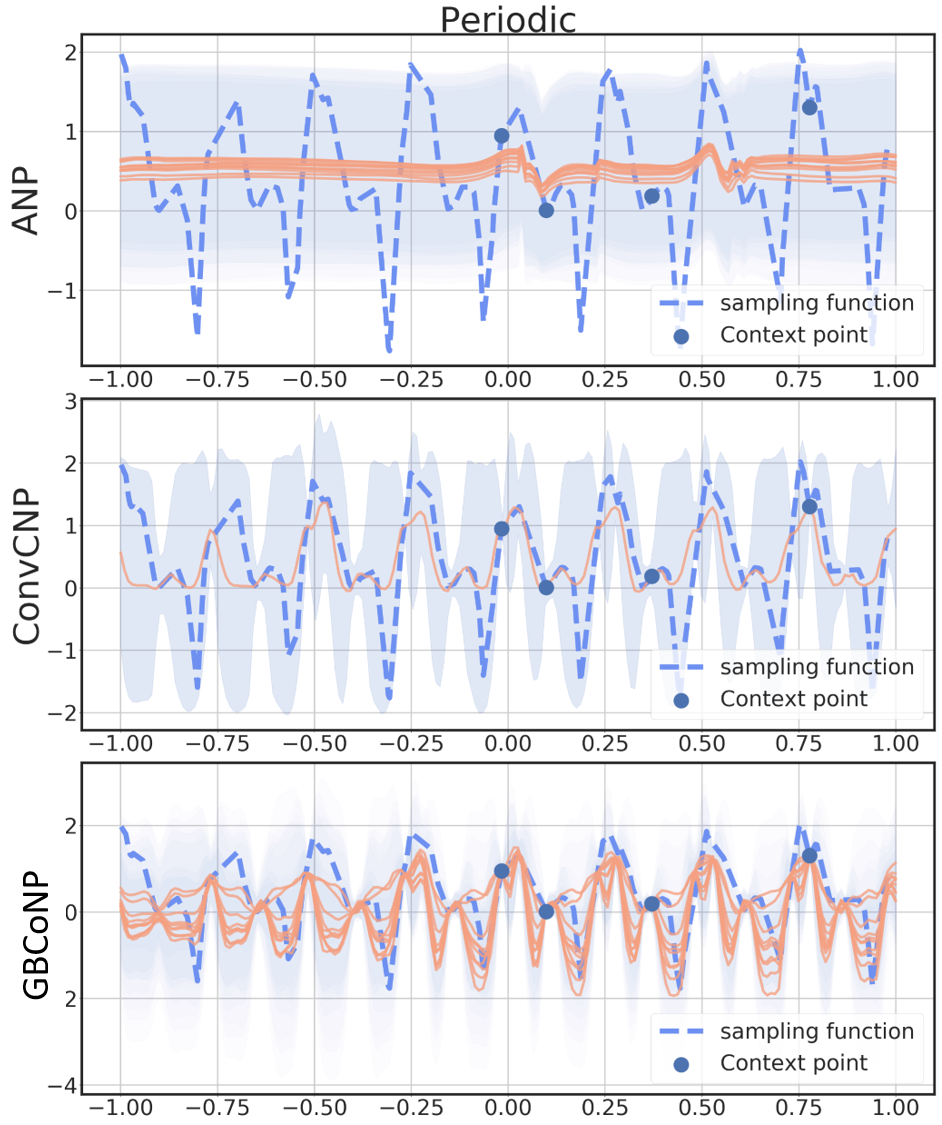}}\\\vspace{5mm}
        {\includegraphics[width=0.32\linewidth]{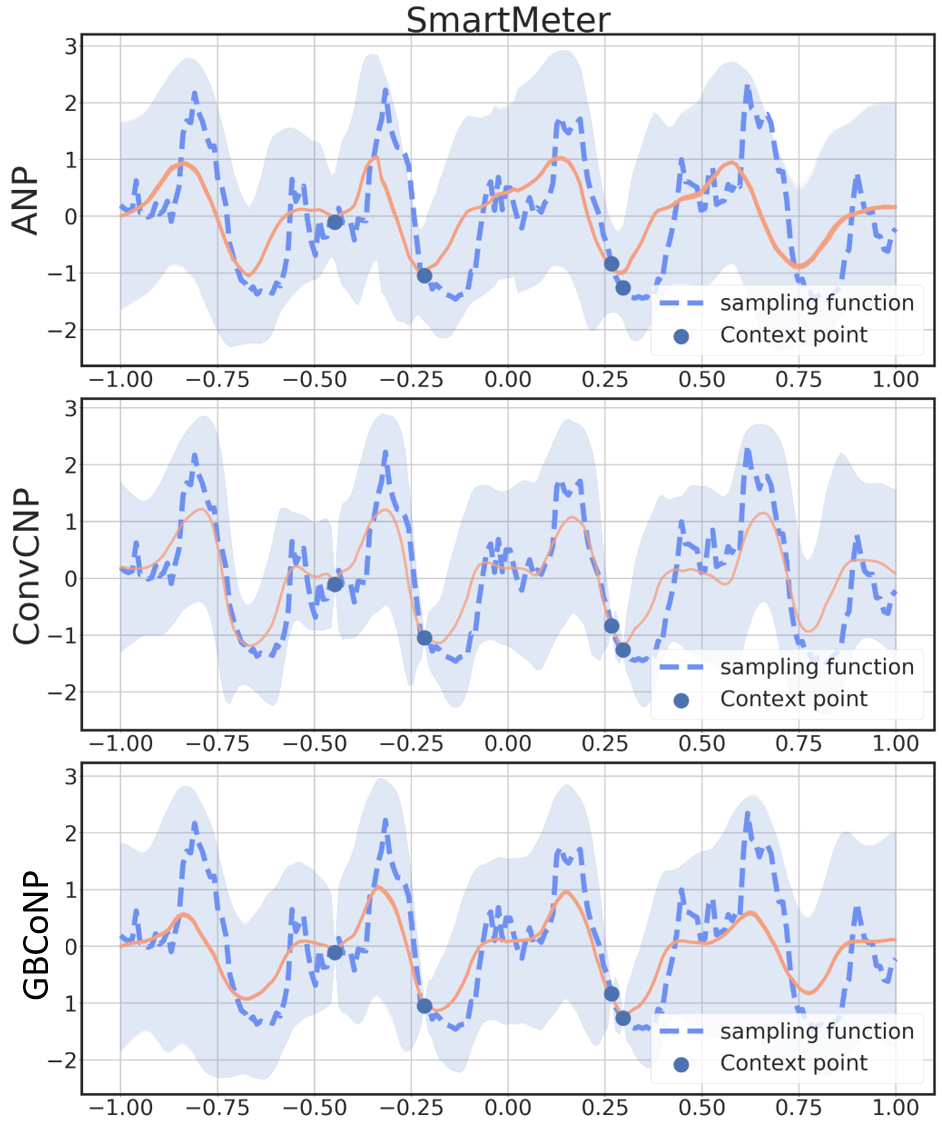}}
        {\includegraphics[width=0.32\linewidth]{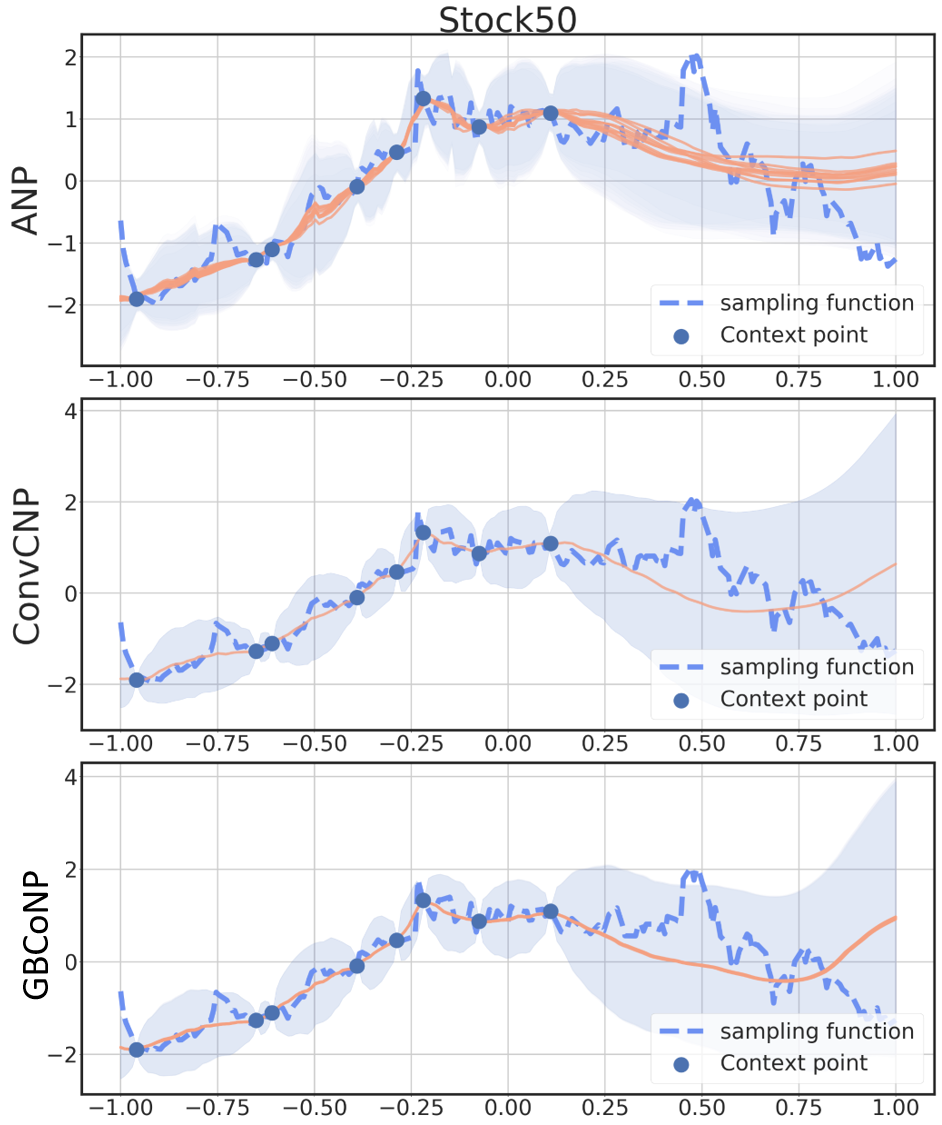}}
        {\includegraphics[width=0.32\linewidth]{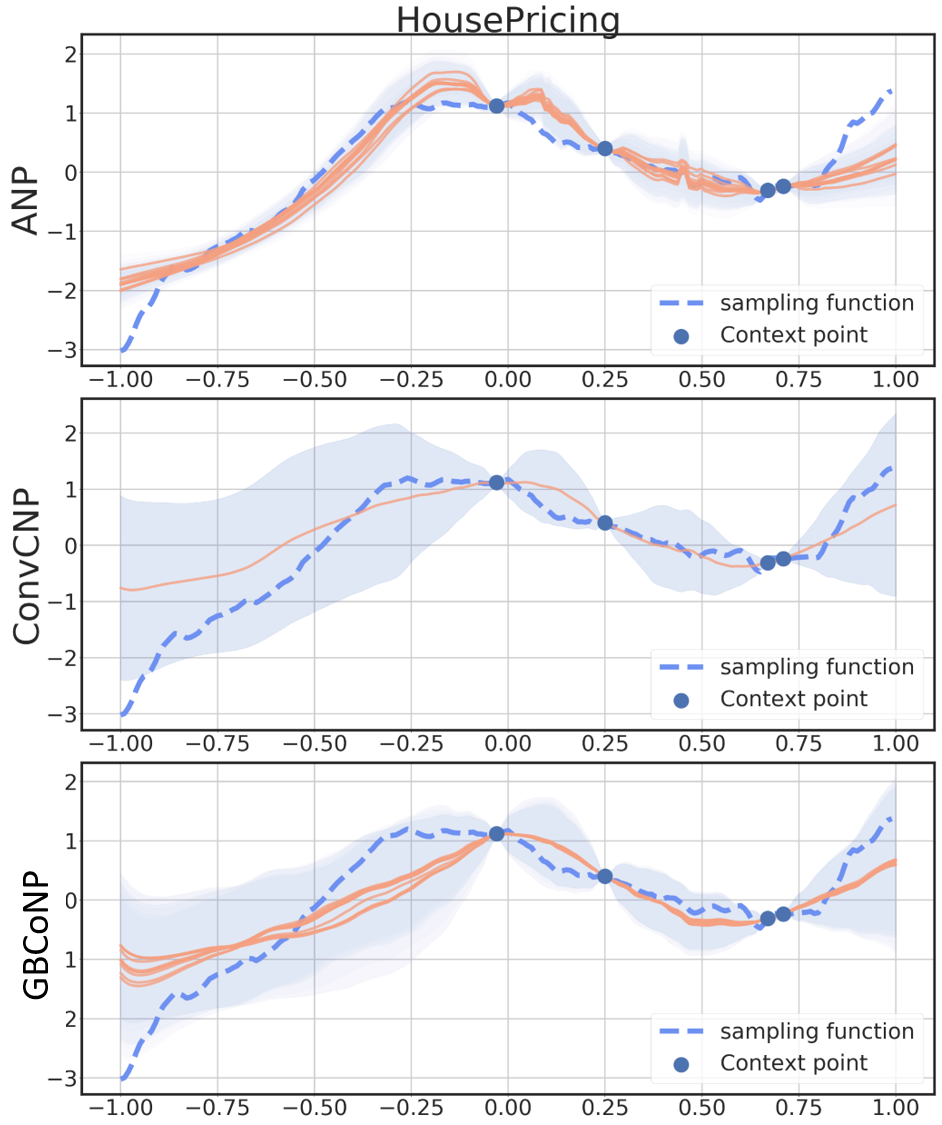}}
    \caption{Model predictions with uncertainty for ANP, ConvCNP and GBCoNP on 6 datasets. For latent based models, 10 different global latent values are sampled and displayed. Each Solid line represents one sample of the predicted mean. The shaded areas, \(\mu \pm 2\sigma\), indicate 95\% confidence intervals. Ground truths are showed in dotted lines, with the context data highlighted in big dots.}
    \label{fig: 1d results}
\end{figure*}

Almost all the ground truths lie in the predicted uncertainties for the three methods (except for the periodic kernel), validating that NP families are capable of modeling uncertainty. The highest uncertainty is normally achieved at the furthest point away from all the context points. The latent NP members generate more diverse samples when compared with their counterpart conditional members at the expense of slight performance drop. This may be attributed to the inference of the intractable latent variables z using context data---the resulting distributional gap \(\mathcal{KL}(q(\textbf{\textit{z}}|\textbf{\textit{r}}_\mathcal{S}')||q(\textbf{\textit{z}}|\textbf{\textit{r}}_\mathcal{S})\) causes information loss in exchange for global uncertainty. 
GBCoNP and ConvCNP produce smoother mean values whereas ANP achieves more diverse samples on real-world datasets.
The most fluctuated predictive means tend to occur in the middle of two context points (see RBF, Matern, and HousePricing in Fig \ref{fig: 1d results}), where the target points are sensitive to both context points. By virtue of their convolution filters, GBCoNP and ConvCNP are able to mitigate those fluctuations.

With regard to global uncertainty \textbf{causal-effect analysis}, we present the values of \(\mu_\textbf{\textit{z}}\) and \(\sigma_\textbf{\textit{z}}\) in \eqref{equ: global uncertainty} after training (Table \ref{tab global uncertainty}). Considering that \(\textbf{\textit{z}}\) is a high dimensional distribution and each task batch presents a pair of \((\mu_\textbf{\textit{z}}, \sigma_\textbf{\textit{z}})\), we therefore averaged the results on both the dimension and batch levels. Table \ref{tab global uncertainty} indicates that the global uncertainty depends largely on model  representation capacity and the intra-task diversity. Model representation capacity refers to constraints a model impose on the global uncertainty. For instance, NP constrains equally loose for all the points, ANP only constrains stricter to the points closer to the context while GBCoNP constrain almost all the points differently with the convolution on \(\mathcal{S}\). As the constraints grow (NP \(<\) ANP \(<\) GBCoNP), the global uncertainty decreases (\(\sigma_\textbf{\textit{z}}\): NP \(>\) ANP \(>\) GBCoNP). Intra-task diversity implies the meta-setting characteristic of the dataset -- how many possibilities there are in this dataset given the same context. For example, although Periodic and SmartMeter are both seasonal datasets, the possible target sets, given a certain context set for Periodic, are not unique; therefore, GBCoNP can predict seasonal means with different amplitudes. In contrast, in SmartMeter, there is only one possible target set corresponding to a context set; causing a smaller \(\sigma_\textbf{\textit{z}}\) in this case( \(<\) Periodic).

\begin{table}[h]
\caption{Global Uncertainty for latent Neural Process Families}
\begin{center}
\begin{tabular}{|c|c|c|c|c|c|c|}
\hline
\textbf{Model}&\multicolumn{2}{|c|}{\textbf{NP}} & \multicolumn{2}{|c|}{\textbf{ANP}} & \multicolumn{2}{|c|}{\textbf{GBCoNP}}  \\
\cline{2-7} 
& \(\mu_\textbf{\textit{z}}\) &  \(\sigma_\textbf{\textit{z}}\) & \(\mu_\textbf{\textit{z}}\) &  \(\sigma_\textbf{\textit{z}}\) & \(\mu_\textbf{\textit{z}}\) &  \(\sigma_\textbf{\textit{z}}\) \\
\textbf{Dataset}$^{\mathrm{a}}$ & (e-3) &   &  (e-3) &   & (e-3) &  \\
\hline 
RBF          & -7.25       & \textbf{0.48}     & -0.48       & 0.20     & 0.29        & 0.19     \\
Periodic    & -1.26       & 0.52     & 7.21        & \textbf{0.57}     & -3.12       & 0.21     \\
Matérn-3/2     & -0.61       & 0.35     & -0.80       & 0.35     & 0.12        & 0.19     \\
Stock50      & 0.74        & 0.29     & 5.93        & \textbf{0.36}     & -0.05       & 0.11     \\
SmartMeter  & -2.67       & \textbf{0.45}     & -0.21       & 0.28     & 10.66       & 0.11     \\
HousePricing & 3.76        & \textbf{0.46}     & 6.58        & 0.25     & 4.53        & 0.18     \\
\hline
MNIST        & -1.00       & 0.32     & 55.01       & \textbf{0.34}     & -1.77       & 0.29     \\
SVHN         & -12.08      & \textbf{0.49}     & -8.61       & 0.13     & 35.76       & 0.11     \\
CelebA32      & 18.30       & \textbf{0.18}     & 2.83        & 0.14     & 50.04       & 0.13     \\
\hline
Covid      & -2.25       & \textbf{0.52}     & -10.56      & 0.13     & -55.88      & 0.44    \\ \hline
\multicolumn{7}{l}{$^{\mathrm{a}}$ \(\epsilon\) in \eqref{equ: global uncertainty} for 1D, 2D datasets are 1, and 5\% total pixels respectively.}\\
\multicolumn{7}{l}{Covid datasets remains providing all previous records as a context set. }
\end{tabular}
\label{tab global uncertainty}
\end{center}
\end{table}

To \textbf{manipulate global uncertainty} for increasing intra-task diversity, we reduce the latent dimension of \(\textbf{\textit{z}}\) from 128 to 4 and display \(\mu_{\textbf{\textit{y}}_\mathcal{T}}\) with different priors in Fig \ref{fig: 1d global uncertainty}. The variance bound in the prior\eqref{equ: global uncertainty} is relaxed  from \(\sigma_z\) to 40 \(\sigma_z\) for Stock50 and Periodic to amplify the effects. The results show that a group of different functions can be sampled meanwhile fitting well with context data. As highlighted in Fig \ref{fig: 1d global uncertainty} (a), some dimension controls the trend of the curve after a context point (up/down) while some others control the amplitude of the trend.
Fig \ref{fig: 1d global uncertainty} (b) shows the controlling factors become the amplitude and the phase of a wave, and the resulting positions of context data in the prediction curves shift from a ``crest" to a ``trough".

\begin{figure}[h!]
    \centering
    \begin{subfigure}[ANP on Stock50]
        {\includegraphics[width=0.98\linewidth]{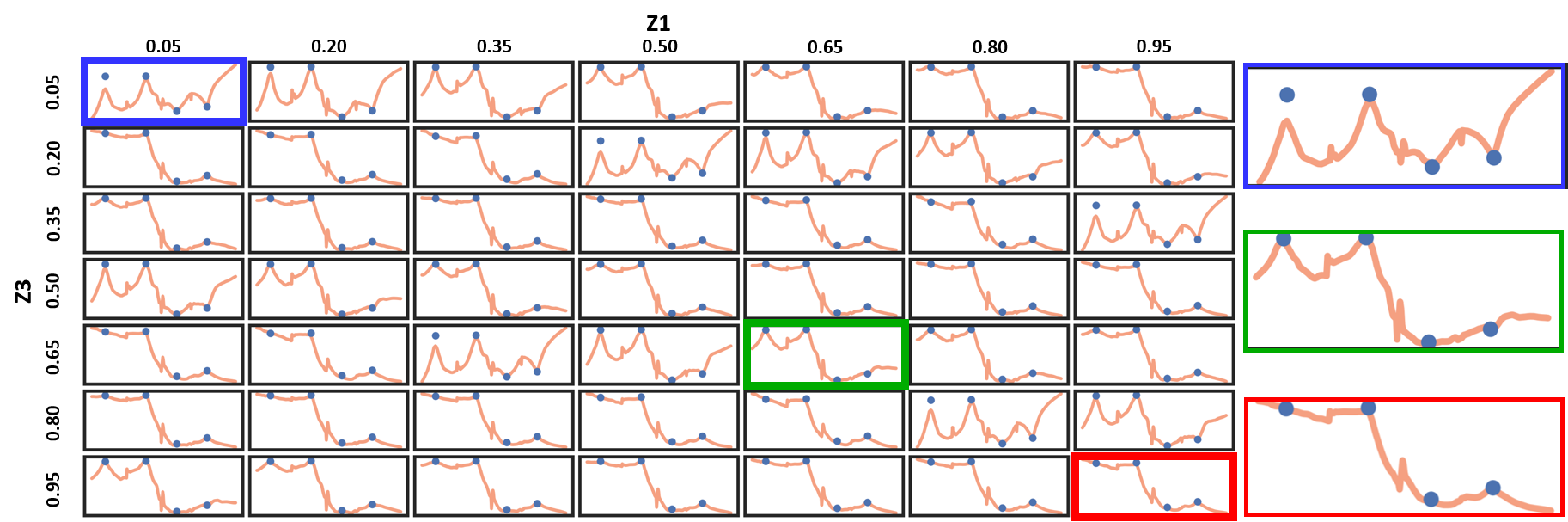}}
    \end{subfigure}
    \begin{subfigure}[GBCoNP on Periodic]
        {\includegraphics[width=0.98\linewidth]{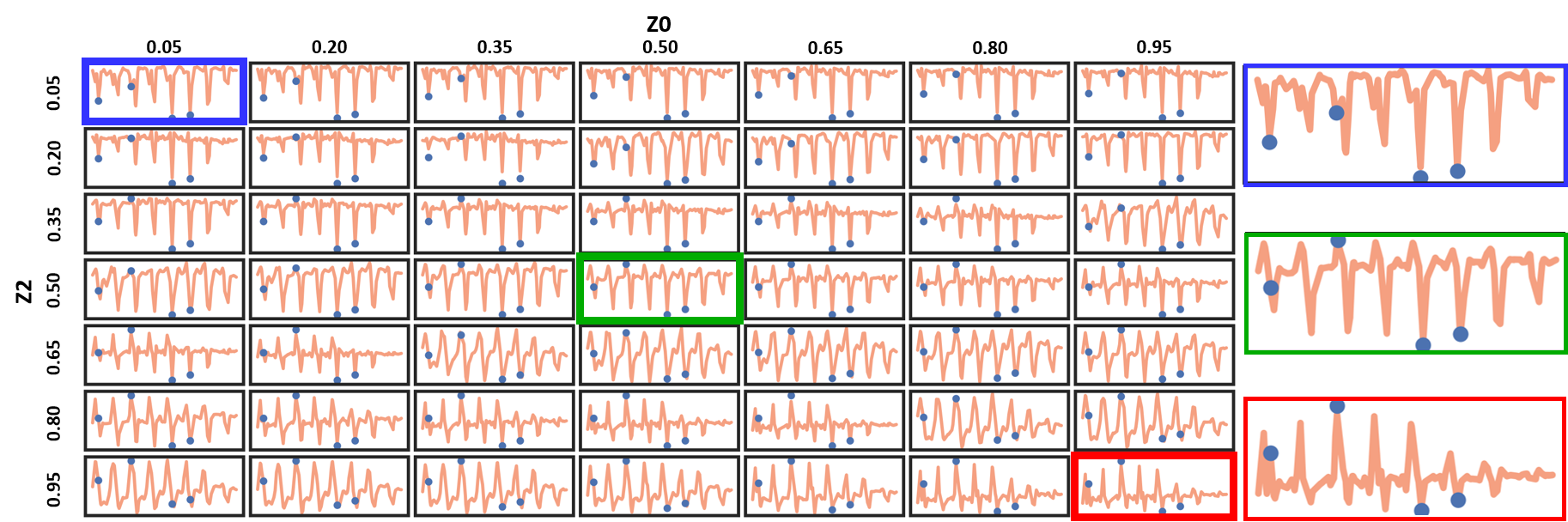}}
    \end{subfigure}
        \caption{Manipulation of global uncertainty for sample generation. A grid of \(7\times7\) latent variables \(\textbf{\textit{z}}\) is sampled to generate predictive means \(\mu_{\textbf{\textit{y}}_\mathcal{T}}\). The sampling distribution \(\mathcal{N}(\mu_z,\sigma^2_z)\) is obtained from the dotted context data. Sampling values vary from the 5\(^{th}\) to the 95\(^{th}\) percentile values.}
        \label{fig: 1d global uncertainty}
\end{figure}

\subsection{2D Datasets}

\(\textbf{MNIST, SVHN, CelebA32}\). In 2D scenarios, we aim to inpaint the whole image with uncertainties given a set of context pixels, and
we select three image datasets (MNIST, SVHN, CelebA32) for the task. For convolutional-based methods, the context values (x) are built using a binary mask \(\in \mathbb{R}^{W \times H}\), where \(\mathcal{U}(0, 0.3(W \times H))\) pixels are unveiled. For non-convolutional baselines, the mask is transformed into a list of 2d inputs representing the relative pixel locations (pixel index/image size). Y values are either 1d or 3d implying the pixel intensity. Each datasets is trained for 50 epochs and tested for 6 runs. Table \ref{tab2d} shows the log-likelihood of the latent and the conditional neural process families on the 2D datasets. GBCoNP achieves the SOTA performance among the latent members. ConvCNP performs the best conditional result on MNIST, and ACNP outperforms on SVHN and CelebA32. 

Fig \ref{fig: 2d results} shows the predictive means and variances on the three datasets. Overall, all the three methods give reasonable results, even with only 5\% of the total pixels. The local uncertainty  occurs on the strokes of a digit for MNIST and SVHN, while in CelebA, the variances lie in  the profile of a face, including face shape, hair, eyes, noise and mouth. The variance on SVHN and CelebA plummets as the amount of context data increases. With 30\% of unveiled pixels, all the models can recover the whole image with little local uncertainty. GBCoNP and ConvCNP can generate smoother predictive means (see Fig \ref{fig: 2d results} MNIST digits ``2" and ``5", and the plate numbers ``42" and ``25" in SVHN). ANP achieves more coherent local uncertainties regarding variances while GBCoNP and ConvCNP tend to reduce the variances around context points to zero.

\begin{table}[h]
\caption{Loglikelihood on 2D and Spatial-Temporal Datasets \\(Mean \(\pm\) Std)}
\begin{center}
\begin{tabular}{|c|c|c|c|c|}
\hline
\textbf{Model}&\multicolumn{4}{|c|}{\textbf{Dataset Name}} \\
\cline{2-5} 
\textbf{Name} & \textbf{\textit{MNIST}}& \textbf{\textit{SVHN}} & \textbf{\textit{CelebA32}}&  \textbf{\textit{Covid}} \\ \hline 
NN      & 1.15 \(\pm\) 0     & 0.06 \(\pm\)  0     & -0.06 \(\pm\)  0   & -3.03  \(\pm\) 0.07 \\
CNP~\cite{garnelo2018conditional}     & 2.03  \(\pm\) 0.03  & 1.26 \(\pm\)  7E-3  & 0.97 \(\pm\)  0.03 & -1.91  \(\pm\) 0.29 \\
ACNP    & 2.04  \(\pm\) 0.03  & \textbf{2.48  \(\pm\) 0.02} & \textbf{1.89  \(\pm\) 0.10} & \textbf{0.53  \(\pm\) 0.58 } \\
ConvCNP & \textbf{2.68  \(\pm\) 0.07}  & 1.84  \(\pm\) 0.28  & 1.87  \(\pm\) 0.11 & 0.34  \(\pm\) 0.41  \\
\cline{1-5}     
NP \cite{garnelo2018neural}  & -8.87  \(\pm\) 2.09 & -1.34  \(\pm\) 0.41 & 0.24  \(\pm\) 0.23 & -1.50  \(\pm\) 0.42 \\
ANP \cite{kim2018attentive}    & 1.36  \(\pm\) 0.19  & 2.48  \(\pm\) 0.03  & 1.66  \(\pm\) 0.16 & 0.67  \(\pm\) 0.69  \\
GBCoNP & \textbf{2.72  \(\pm\) 0.06}  & \textbf{2.57  \(\pm\) 0.03 } & \textbf{1.82  \(\pm\) 0.15} & \textbf{0.83  \(\pm\) 0.10 } \\ \hline
\end{tabular}
\label{tab2d}
\end{center}
\end{table}

\begin{figure}[h!]
    \centering
        {\includegraphics[width=0.96\linewidth]{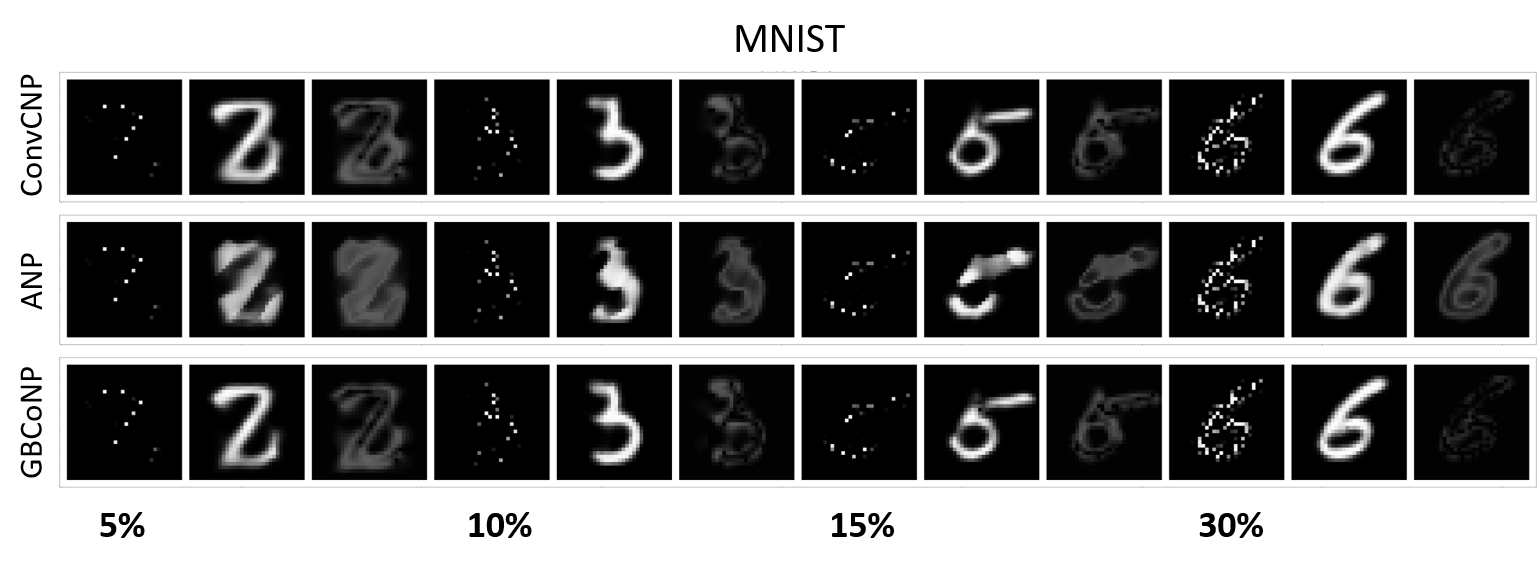}}
        {\includegraphics[width=0.96\linewidth]{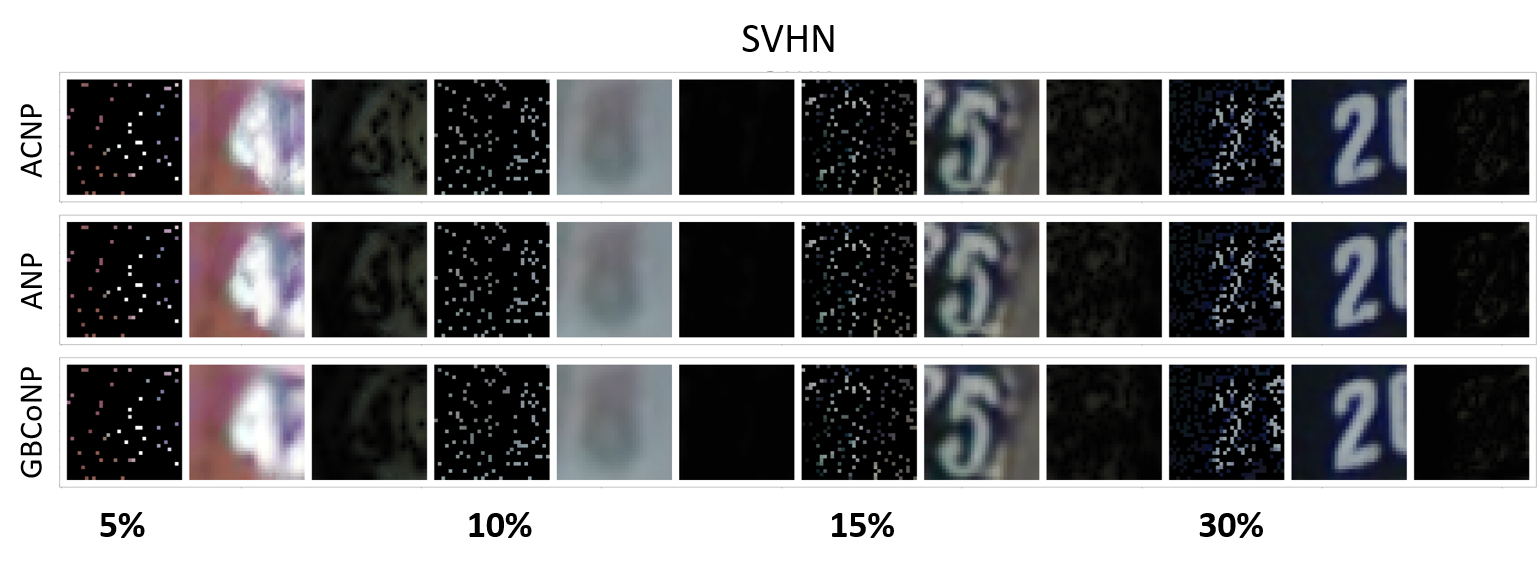}}
        {\includegraphics[width=0.96\linewidth]{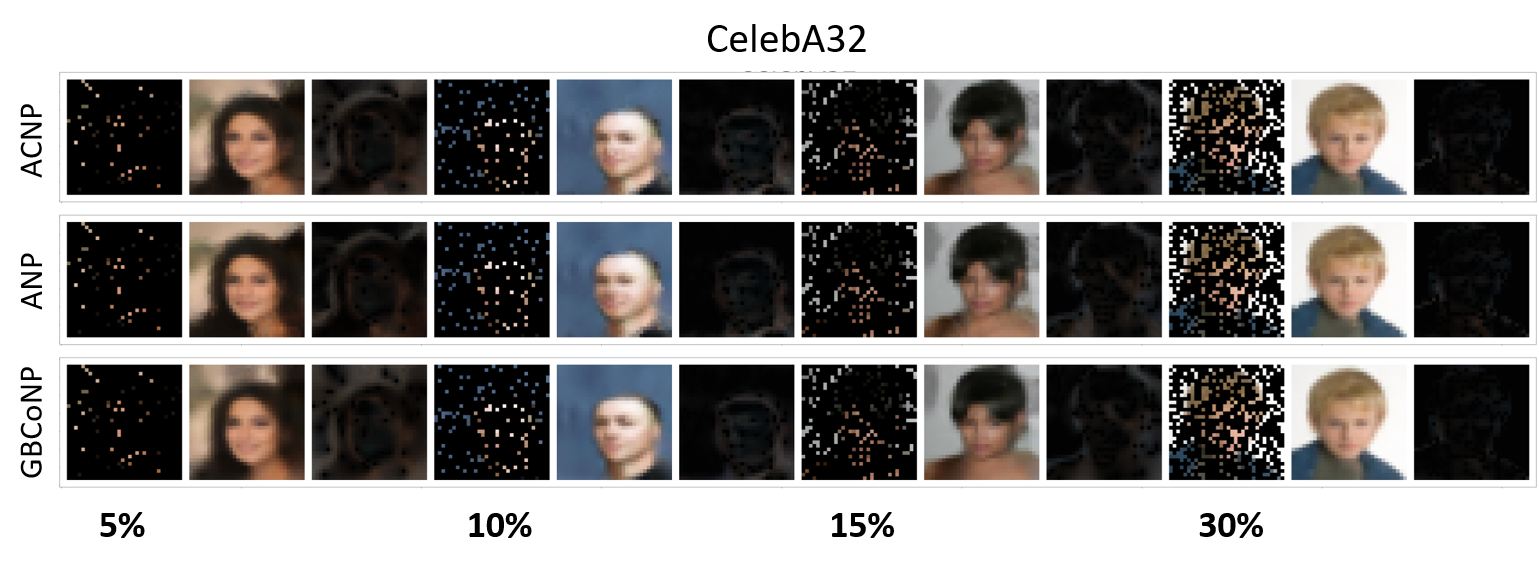}}
    \caption{Model predictions with uncertainties on MNIST, SVHN and CelebA32. For each task, four samples are presented with differed proportions of the unveiled context pixels \(\in\{5\%, 10\%, 15\%, 30\%\}\). Each sample gives three consecutive images: masked images, predictive means, and standard deviations.}
    \label{fig: 2d results}
\end{figure}

As shown in Table \ref{tab global uncertainty}, when the intra-task diversity is large enough, the global uncertainties achieved by several methods are quite similar (NP \(\approx\) ANP  \(\approx\) GBCoNP in MNIST). For instance, given a small set of pixels indicating a number ``6", there are plenty of possibilities in this dataset to finish the inpainting, therefore producing a large uncertainty \(\sigma_z\).To manipulate this global uncertainty, we generate images with respect to different priors in Fig \ref{fig: 2d global uncertainty}. Similar to 1D, we relax the prior standard deviation to 12 \(\sigma_z\). For MNIST, the global uncertainty in ANP determines the thickness and extension of a stroke, e.g., a ``1" can be transformed to a ``9" then to a ``4" if the upper stroke of ``1" is gradually extended and thickened.
Besides, extension towards different directions can result in variants of orientations of a digit (see digit ``6" in Fig \ref{fig: 2d global uncertainty}). The results of GBCoNP comply with this extension pattern yet with limited variation due to the model representation capacity as discussed in 1D. For CelebA32, global uncertainty depicts the background and appearance. Changing values across the prior can transit the background color from completely black to white with different shades of brightness. Appearance variations range from the size of the eyes, noses, face shape, to the hair volume and color.

\begin{figure*}[h!]
    \centering
        {\includegraphics[width=0.24\linewidth]{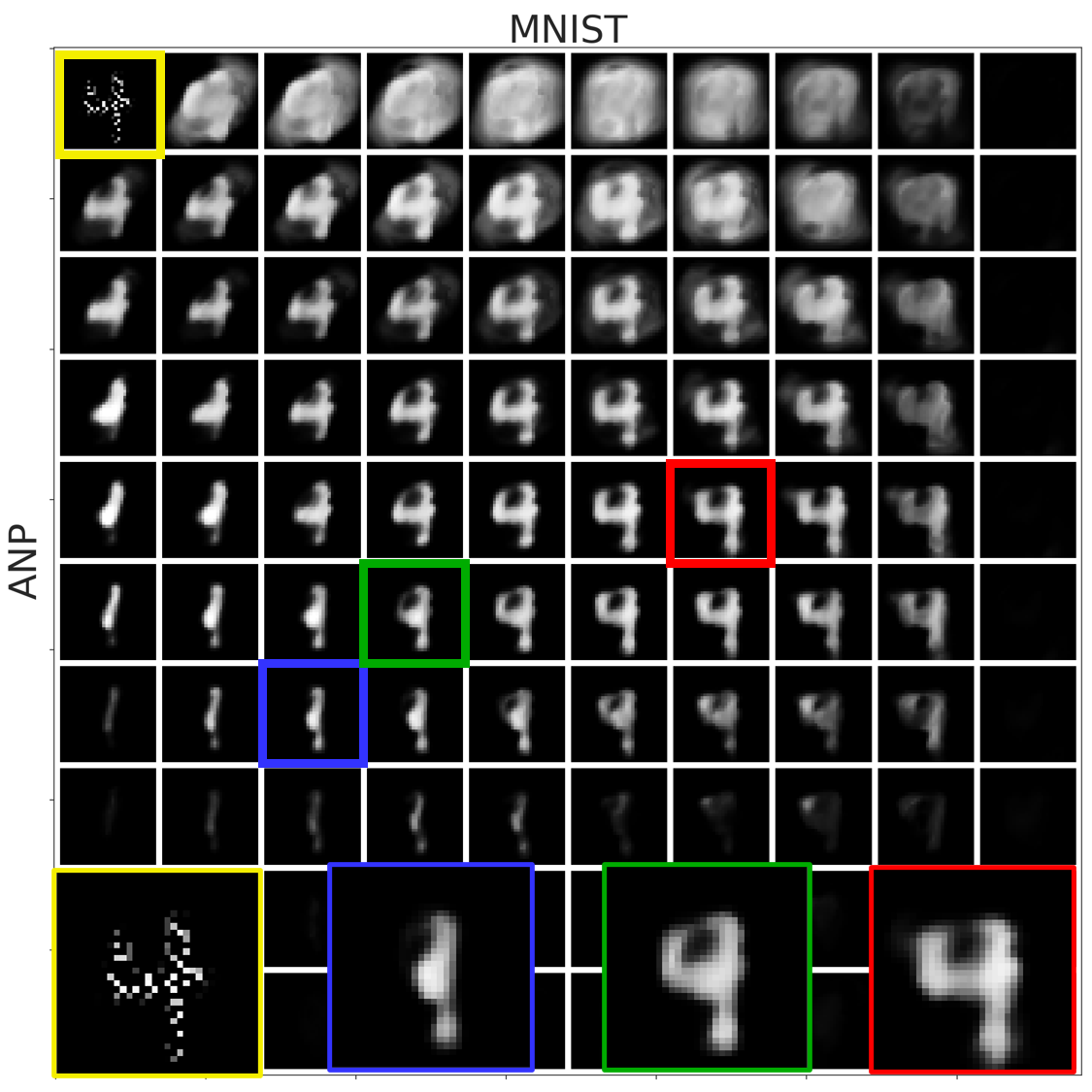}}
        {\includegraphics[width=0.24\linewidth]{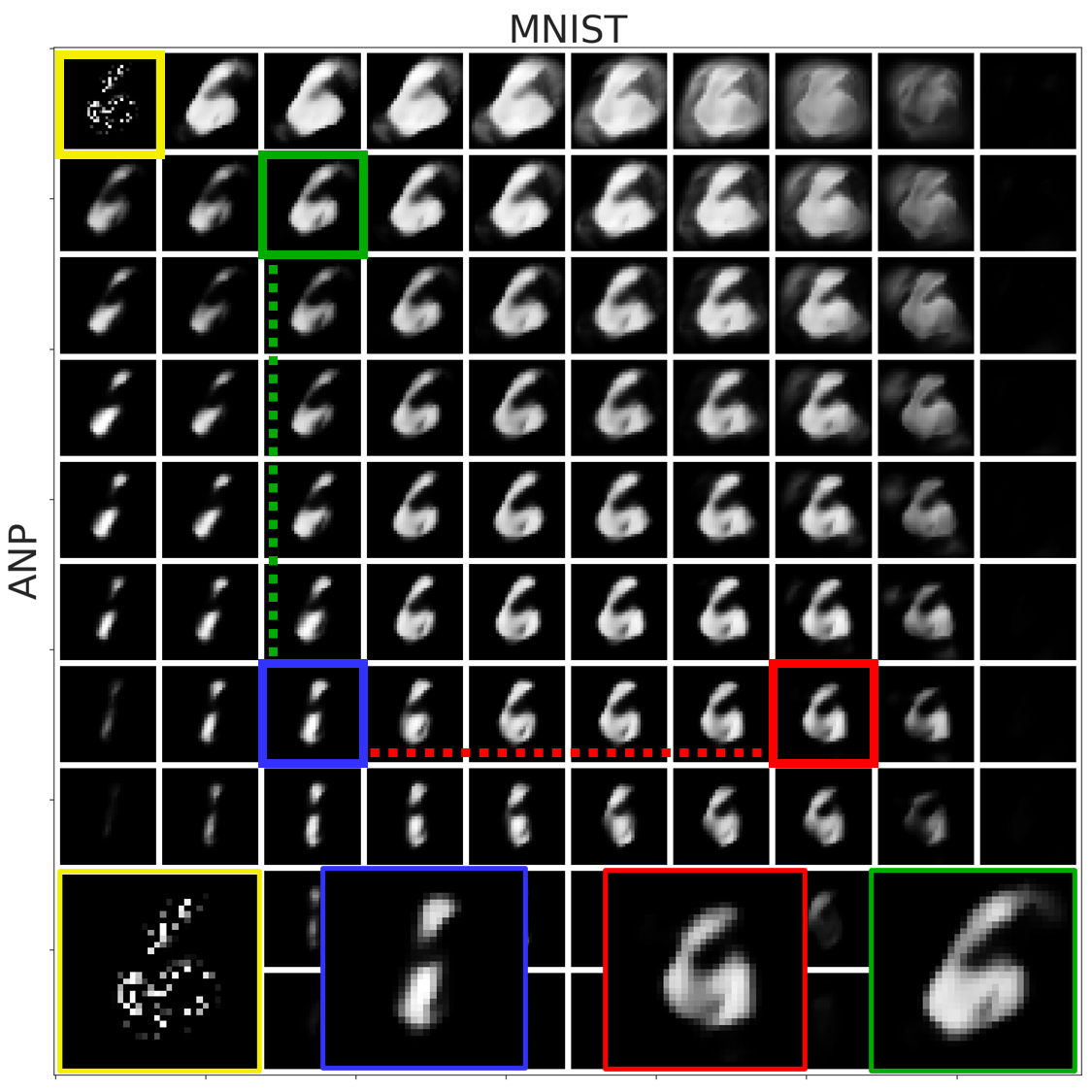}}
        {\includegraphics[width=0.24\linewidth]{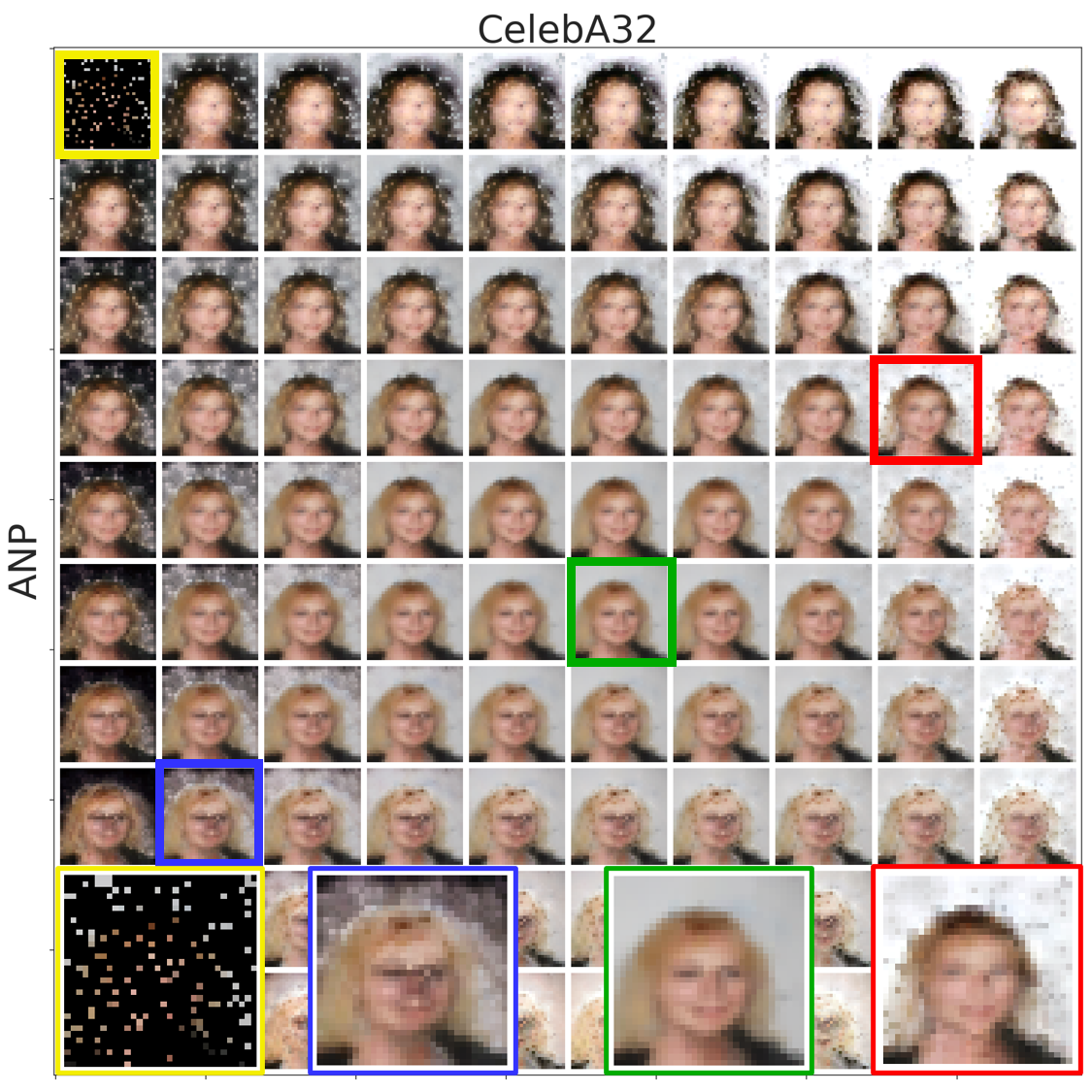}}
        {\includegraphics[width=0.24\linewidth]{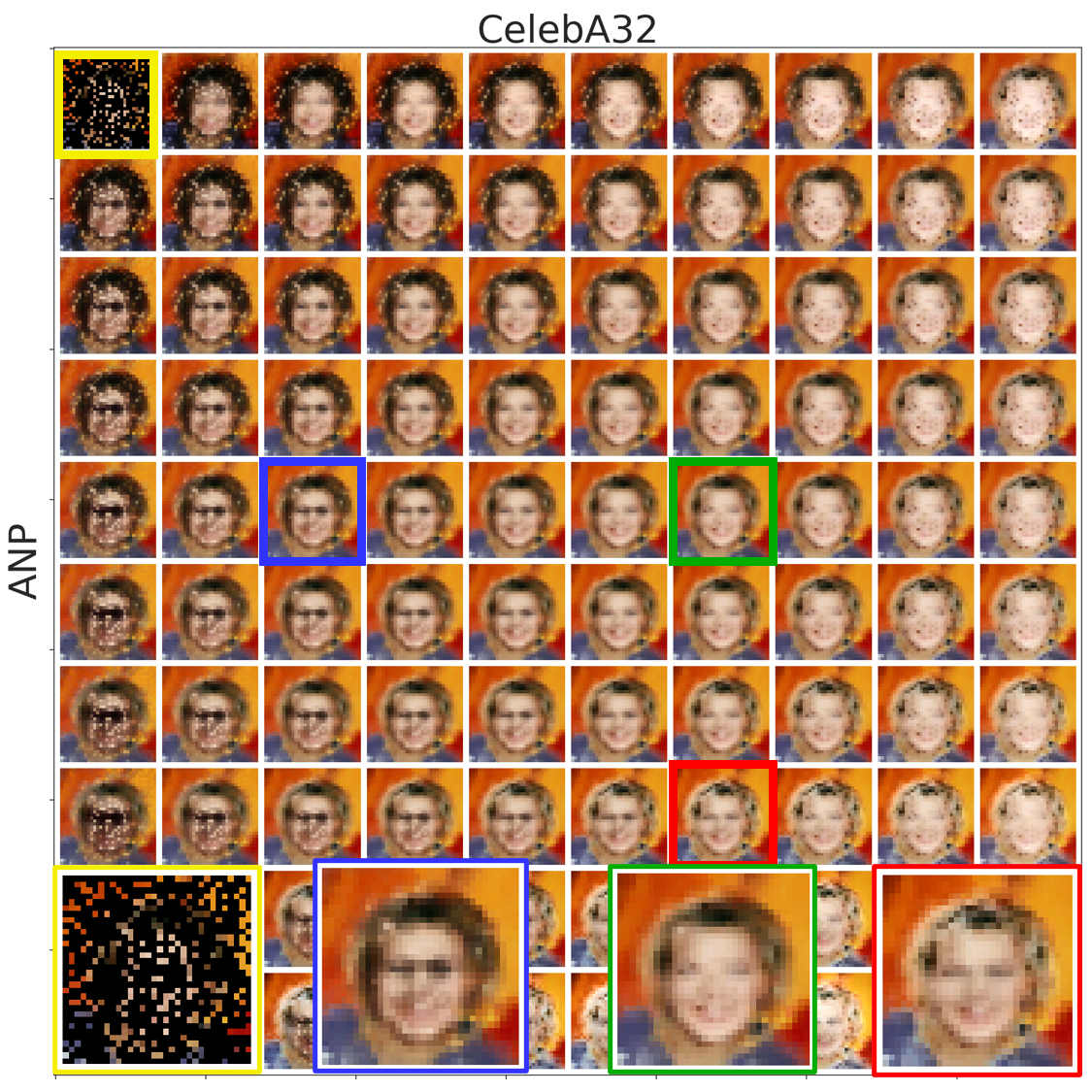}} 
        \vspace{3mm}
        {\includegraphics[width=0.24\linewidth]{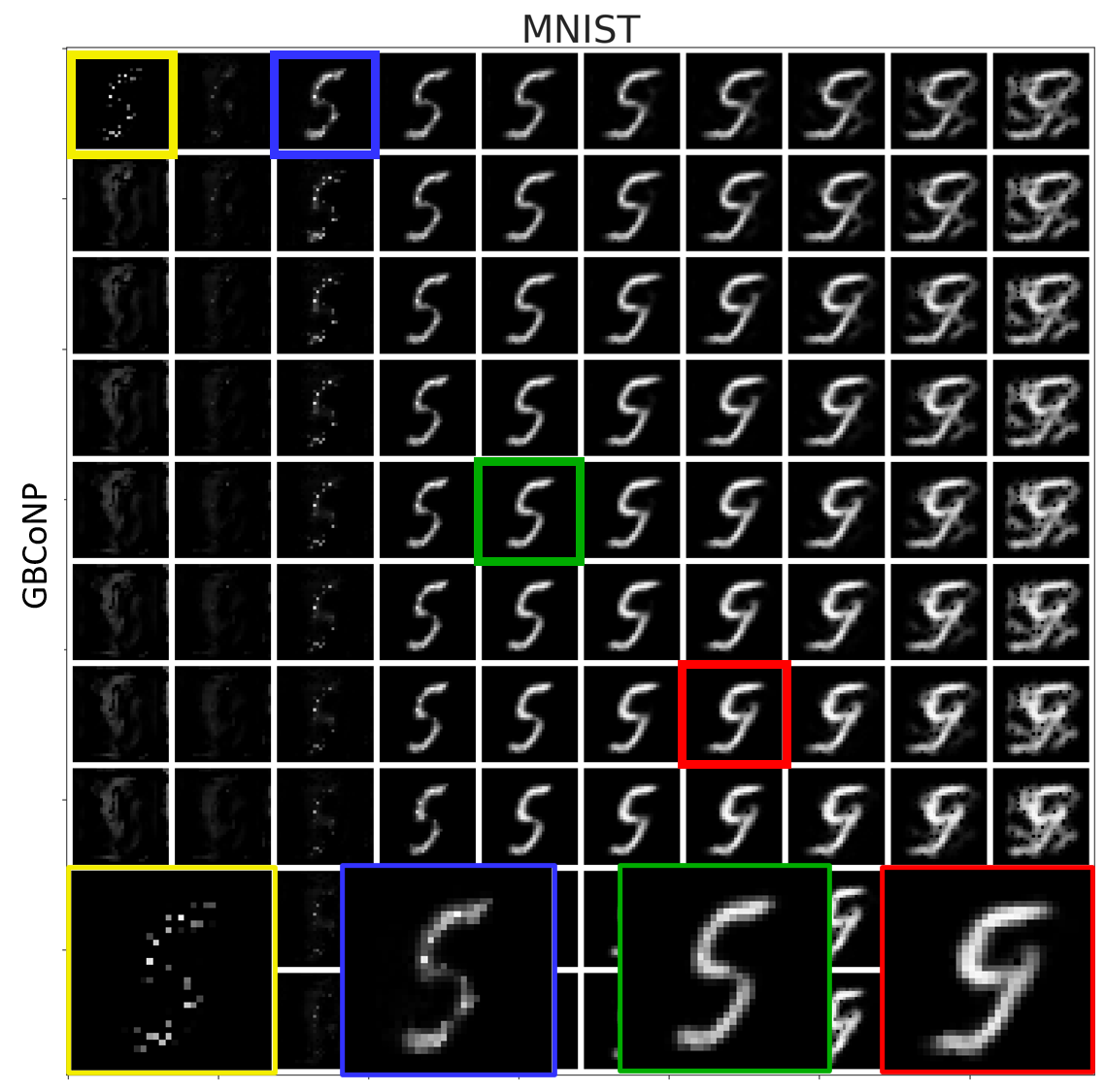}}
        {\includegraphics[width=0.24\linewidth]{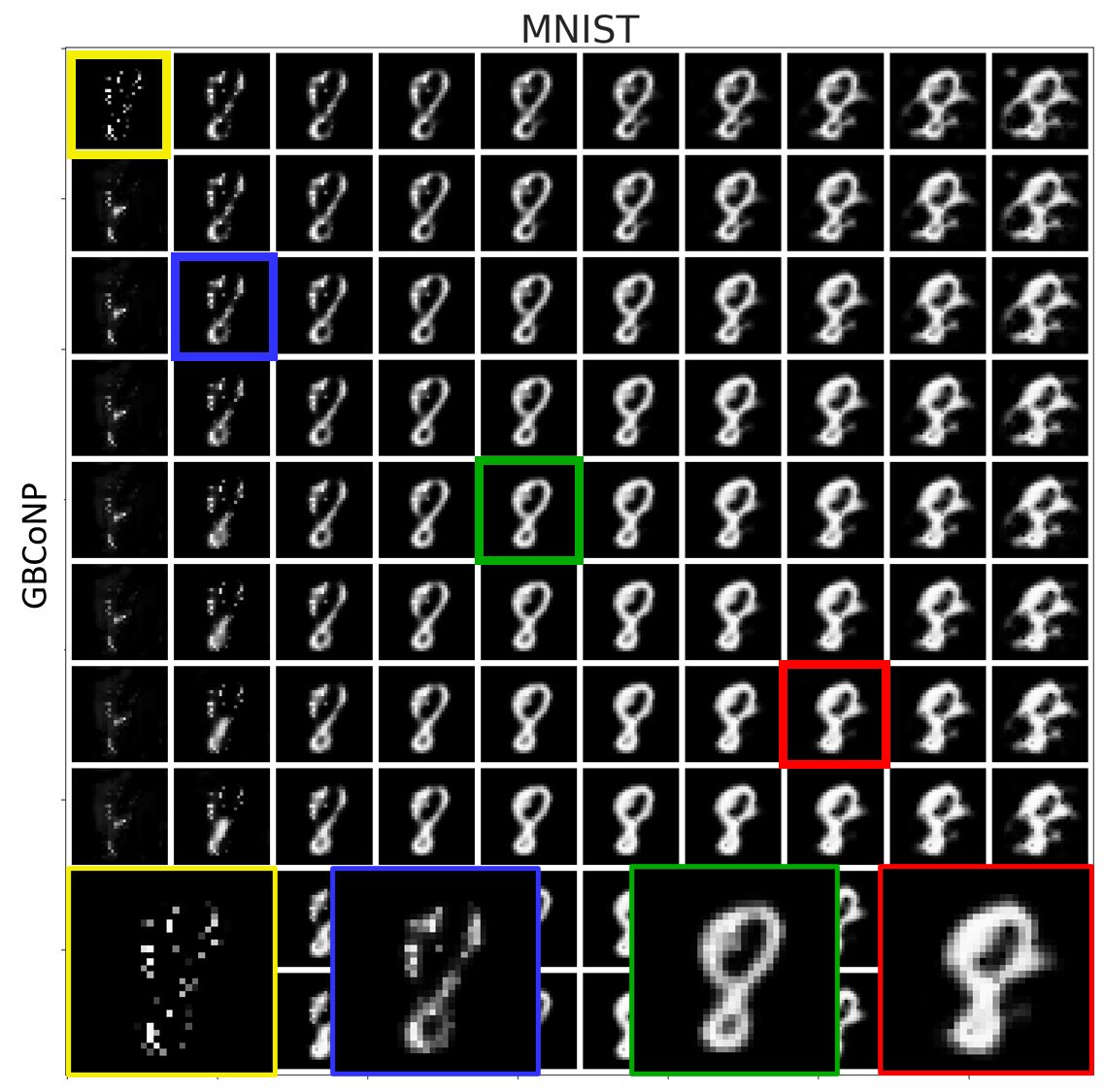}}
        {\includegraphics[width=0.24\linewidth]{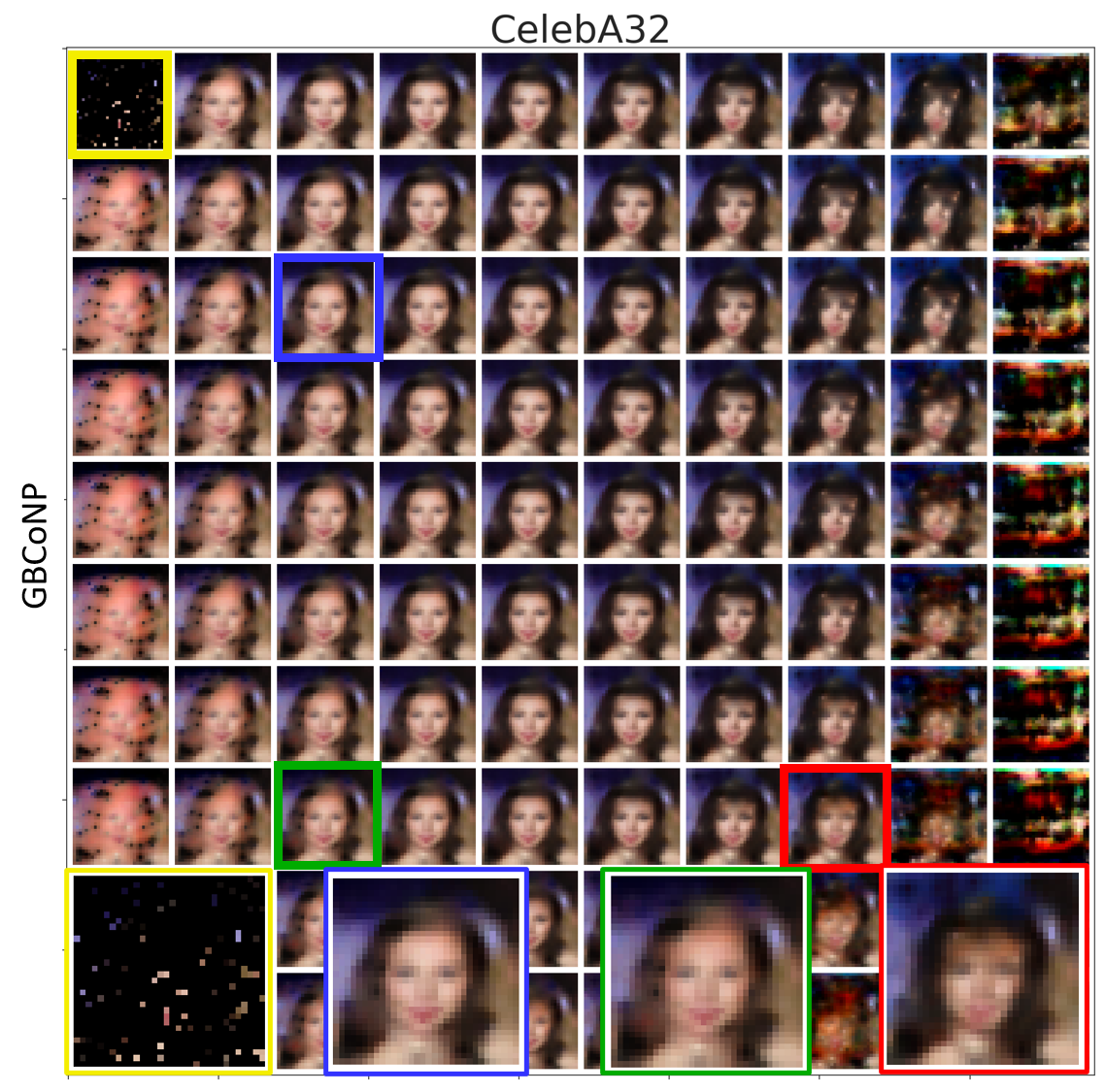}}
        {\includegraphics[width=0.24\linewidth]{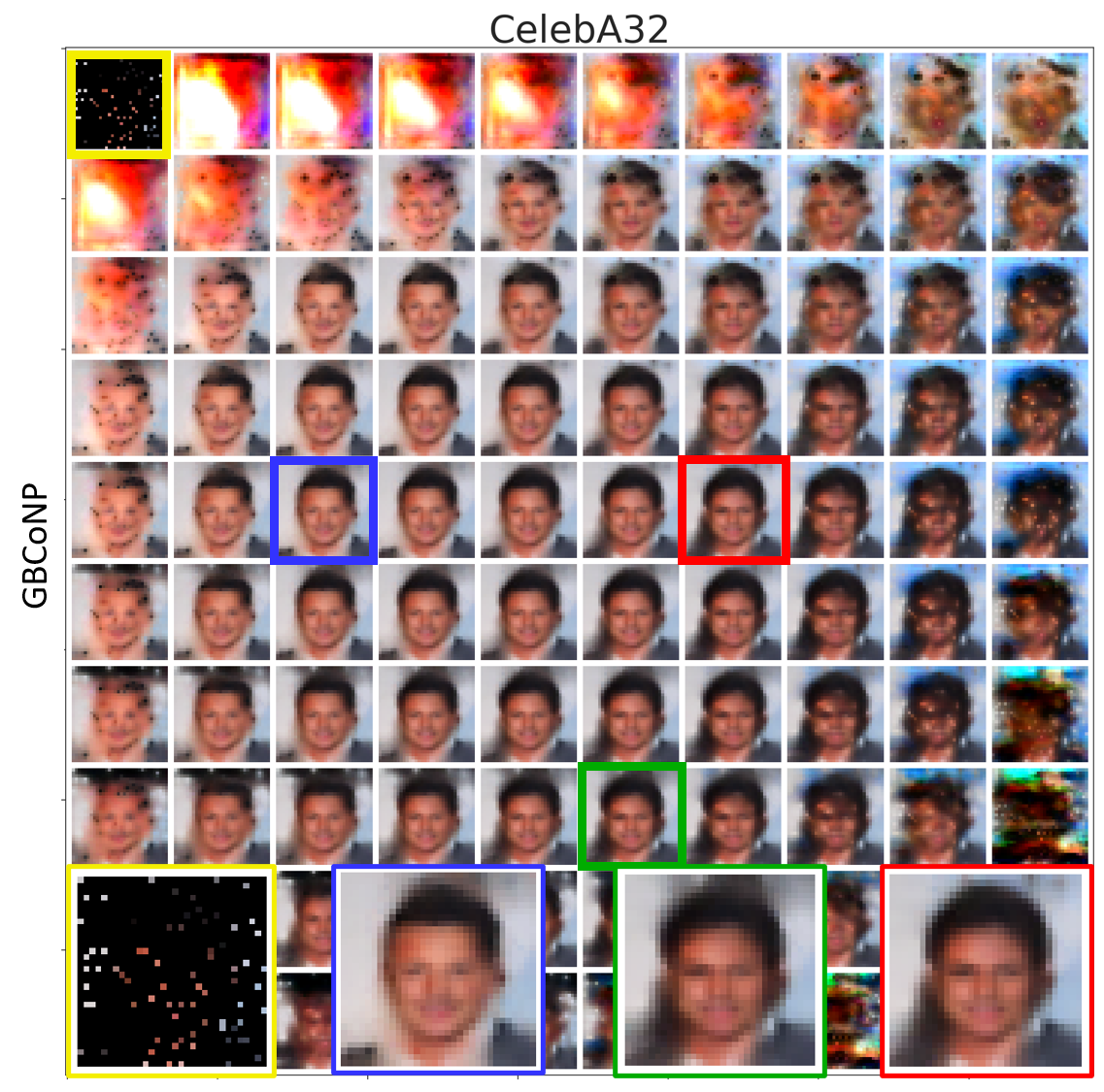}}
    \caption{Manipulation of the global uncertainty for ANP and GBCoNP on MNIST and CelebA32. A grid of \(10\times10\) different latent variables z is sampled to generate predictive means. Four bounding boxes in the grid are enlarged at the bottom line for clarity. The first element is the masked image, i.e., context data. The other three are diverse samplings for comparison.}
    \label{fig: 2d global uncertainty}
\end{figure*}

\subsection{Spatial-Temporal Dataset} 
\(\textbf{COVID}\)\footnote{\url{https://www.kaggle.com/fireballbyedimyrnmom/us-counties-covid-19-dataset}}. In spatial temporal scenarios, we use this dataset that contains daily total confirmed coronavirus cases in all counties of US from 21/01/2020 to 20/04/2021. We use a window of 14 days with daily total confirmed cases sampled in each task to build temporal features that could depict the trend. Each element is normalized by \(\log{(x - x_{min} +1)}\), where \(x_{min}\) suggests the lowest cases therein.
Since three points are sufficient to describe the curve, i.e., the last 7 days, the last 3 days and today, according to our observation, these context points are used for predicting the logged relative growth 7 days after. To fully utilize spatial information, the data for each county are projected on a geographic map (shown in the first row Fig \ref{fig: spatial-temporal result}), where spatial relationships between counties are displayed while preserving temporal information with color intensity.

The whole map is segmented into a grid of 104 of 40\(\times\)40 cells and a model only processes one cell at a time to minimize the computational cost. For convolutional methods, context values x is a binary mask \(\in \mathbb{R}^{T\times W \times H}\), where only the last time channel is set to zero. For non-convolutional baselines, the mask is transformed into a list of 3d inputs with the relative spatial and temporal locations where Y values correspond to the log-relative growth. The dataset is trained for 100 epochs and tested for 6 runs.

The last column of table \ref{tab2d} compares the log-likelihood metrics among NP families. GBCoNP and ACNP achieve the SOTA performance in their categories, and GBCoNP outperforms ConvCNP in stability. Most models have high log-likelihood variances due to random sampling of the cell. In contrast with non-convolutional neural processes that could only handle a maximum of \(40 \times 40 \times 3 = 4800\) (one cell in the grid) context points each time, GBCoNP is the only one that can process the whole grid (104 cells) by convolutions which results in a more stabilized metric. Interestingly, the locations with the low mean values tend to have high uncertainty (regions with whiter means have redder variances), as shown in Fig \ref{fig: spatial-temporal result}(b) and (c)---the model believes that these ``safer" regions surrounded by high risk neighbourhoods have higher tendency to be infected in the future. Besides, GBCoNP yileds more precise mean values (purple line in Fig \ref{fig: spatial-temporal result}(c)) with higher standard deviations compared with ANP (see Fig \ref{fig: spatial-temporal result}(b)).
Our supplementary experiments on global uncertainty reveal that the COVID dataset has the highest intra-task diversity compared with other datasets, as this covid spread trend has the most diverse possibilities. Manipulation on different priors show the standard deviation change of a region from fairly low risk to high risk.

\begin{figure*}[h!]
    \centering
    \begin{subfigure}[Context and target ground truth]
        {\includegraphics[width=0.9\linewidth]{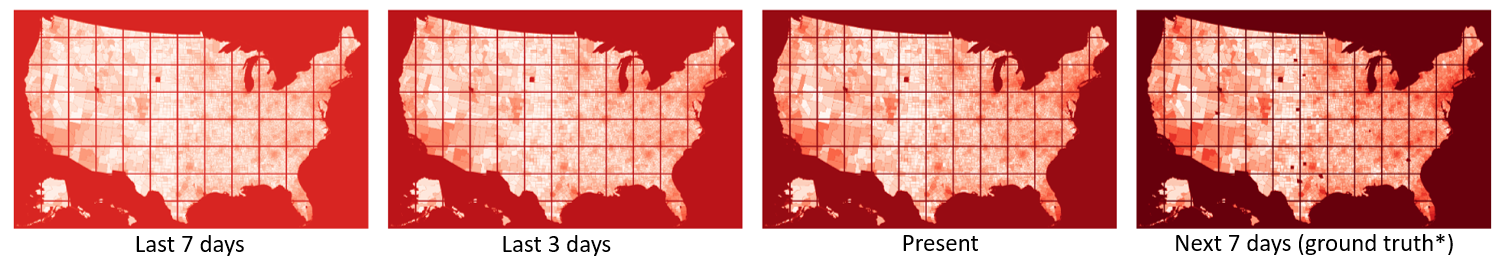}}
    \end{subfigure}
    \begin{subfigure}[Spatial predictions with uncertainties]
        {\includegraphics[width=0.9\linewidth]{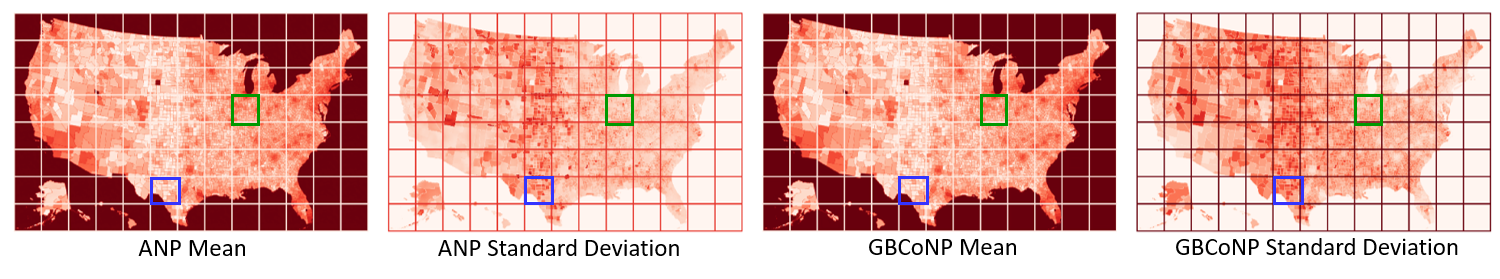}}
    \end{subfigure}\begin{subfigure}[Temporal predictions with uncertainties]
        {\includegraphics[width=0.85\linewidth]{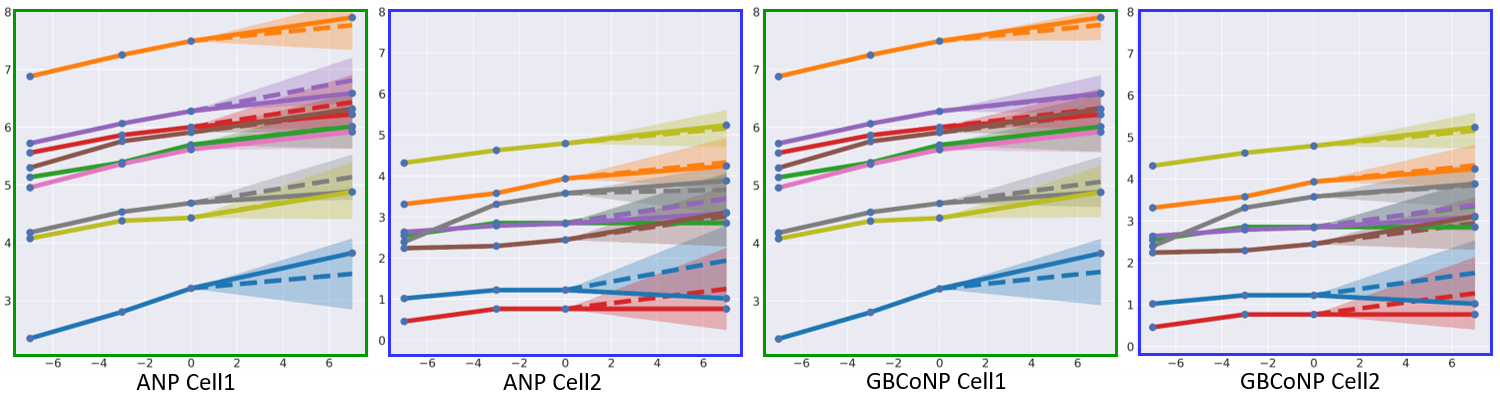}}
    \end{subfigure}
    \caption{Model predictions with uncertainties on COVID (sample date: 01-April-21). (a): Context data include the last 7, 3 days and the present day. The shades of color represent the relative growth compared with the data in the last 14 days. Saturation of the colors in context data are adjusted in the plot to increase contrast for better visualization. (b) Spatial predictions with uncertainties for ANP and GBCoNP. Each model processes one cell in the grid at a time. (c) Temporal predictions on two \(40 \times 40\) cells. Nine pixels are presented in each cell. The ground truth are depicted as solid lines. Predictions are shown in dotted lines with the shaded area indicating uncertainty.}
    \label{fig: spatial-temporal result}
\end{figure*}

\section{Related Works}
\textbf{Neural Processes.} To enhance uncertainty modelling with Gaussian Process(GPs), previous efforts~\cite{garriga2018deepnnasgps}\cite{wilson2016deepkernel} focus on representing deep kernels in GPs before Neural Processes(NPs) were proposed. Although they inherited the powerful feature representation from neural networks(NNs), they still suffer from computationally expensive matrix inversion \(\mathcal{O}(n^3)\). NPs novelly represent stochastic processes with neural netorks under two constraints: exchangeability and consistency, which frees the model from inversion with fully back-propagation and induces the whole neural processes family \cite{dubois2020npf}. There are two major branches in the family: conditional NPs(CNP)~\cite{garnelo2018conditional} and latent NPs. Others members add different inductive biases on context and target relationships; some famous ones include Attentive NPs(ANPs)~\cite{kim2018attentive}, Sequential NPs(SNPs)~\cite{singh2019sequential}, and Convolutional CNPs (ConvCNPs)~\cite{garnelo2018conditional}. A recent work, ConvNP~\cite{foong2020meta_convnp}, adds latent path to ConvCNP; but instead of using variational inference, it adopts Monte Carlo estimation for z and focuses on local coherence. Inspired by this, our work further elaborates on the global uncertainty and the related effect.

\textbf{Generative model with priors.} Generative models help increase intra-diversity of data distributions and mitigate few shot learning issues. They generally generate sythetic samples with priors over model condition. For deep learning-based time series generation, \cite{tang2020joint1dgenerration} applies LSTMs to capture context information and fill the missing values. \cite{yoon2019time} uses adversarial nets to generate sequence with temporal dynamics across time. In image inpainting with priors, conditional VAEs~\cite{ulyanov2018deepiamgeprior} and GANs~\cite{gu2020ganprior}, \cite{shen2020interpretingganprior} manipulate the semantic features in an image (e.g., "smile", "eyeglasses", "age"). However, very few existing studies have meta-learning settings for a distribution of sampling functions, even fewer have considered model uncertainty. \cite{xu2020maml-gp} employs a meta-agnostic structure for reinforcement learning but uses GPs rather than NPs.

\section{Conclusion}
In this paper, we answered three important questions about the global uncertainty in latent neural processes: How to formalize global uncertainty? What causes and affects global uncertainty? How to manipulate the global uncertainty for data generation? We define the global uncertainty as a prior of \(\textbf{\textit{z}}\) from a latent functional distribution given a small set of context data. We discovered that global uncertainty is affected by the model representation capacity and the data intra-task diversity. Our manipulation of the global uncertainty not only achieves generating the desired samples to tackle few-shot learning, but also enables the probability evaluation on the functional priors.

\begin{appendices}
\section{Dataset details}
\label{appendix:Dataset details}
For 1d datasets, We use synthetic Gaussian processes with 3 kernels to sample values. The kernel functions are as follows:
\begin{itemize}
    \item \(\textbf{RBF}: \quad \mathcal{K}(x, x^\prime) = e^{-\frac{1}{2}(\frac{x - x^\prime}{0.2})^2}\)
    \item \(\textbf{Periodic}: \mathcal{K}(x, x^\prime) = e^{-2(\frac{\sin{2\pi(|x - x^\prime
    |)}}{0.5})^2}\)
    \item \textbf{Matern\(-\frac{3}{2}\)}: \quad \(\mathcal{K}(x, x^\prime)=(1+5\sqrt{3}|x - x^\prime|)e^{-5\sqrt{3}|x - x^\prime|}\)
\end{itemize}
where the x values lie within \(x\in\) [-2, 2], and the function values y are sampled from \(y \sim \mathcal{GP}(0, \mathcal{K}(x, x^\prime))\). Training, validating, and testing data sizes are set to 50,000, 10,000 and 5,000.

\textbf{Stock50}\footnote{\url{https://www.kaggle.com/rohanrao/nifty50-stock-market-data}} contains daily trading volumes of 50 stocks from National Stock Exchange India. In each sampling task, 200 days of data are sampled starting from a random date between 11/2016 to 11/2017. X values represent the days count from the first day, and y values represent the corresponding volume weighted average price. Training, validating and testing sizes of the stocks are 36, 10, and 4, respectively. 

\textbf{SmartMeter}\footnote{\url{https://www.kaggle.com/jeanmidev/smart-meters-in-london/version/11}} includes half-hourly average energy consumption readings from 5,567 London households during 03/12/2011 -- 28/02/2014~\cite{3springs2019github}. For each task, 100 hours of data are sampled with a random timestamp. X values refer to the relative time gaps measured in days, and y values refer to the consumption in kWh/half-hour. The proportions of the training, validating and testing sizes are 7:2:1 based on time line.

\textbf{HousePricing} \footnote{\url{https://www.kaggle.com/paultimothymooney/zillow-house-price-data?select=City_Zhvi_AllHomes.csv}} comprises monthly average house prices in 9,473 American cities from 01/1996 to 03/2020. One hundred months of data are sampled per sampling task. X values are the time differences measured in months, and values y are prices in dollars. The proportions of the training, validating and testing sets are 7:2:1 based on total number of cities index.

\section{Training details}
\label{appendix: running time costs}
The running time costs of GBCoNP and ConvCNP are compared in Table \ref{tab running time costs}. The total number of epochs for 1D, 2D and Covid datasets is 100, 50, 100 respectively. 
\begin{table}[h]
\caption{Wall-clock Time costs in Seconds \\ (Mean \(\pm\) Std) on 5 Epochs}
\begin{center}
\begin{tabular}{|c|c|c|c|c|}
\hline
\textbf{Model}&\multicolumn{2}{|c|}{\textbf{GBCoNP}} & \multicolumn{2}{|c|}{\textbf{ConvCNP}} \\
\cline{2-5} 
\textbf{Dataset} & mean & std & mean & std\\
\hline 
RBF          & 17.53      & 0.11   & 7.60 & 0.06   \\
Periodic    & 17.59     & 0.07   & 7.48 &  0.11     \\
Matérn-3/2     & 17.54       & 0.10  & 7.49 & 0.14  \\
Stock50      & 33.42        & 0.14  & 20.86 &  0.14   \\
SmartMeter  & 50.87       & 0.10  & 38.49 &  0.41 \\
HousePricing & 41.61    & 0.09  & 21.97 & 0.24 \\
\hline
MNIST        & 618.30       & 6.04  & 188.60 & 0.78 \\
SVHN         & 763.44      &  7.14  & 234.04 & 0.36  \\
CelebA32      & 2322.96   &  3.53  & 854.94 & 2.32 \\
\hline
Covid      & 93.46    & 3.53  & 82.05 & 0.18\\ \hline
\end{tabular}
\label{tab running time costs}
\end{center}
\end{table}
\end{appendices}

\bibliographystyle{IEEEtran}
\bibliography{bio}

\end{document}